\pdfoutput=1
\documentclass{article} 
\usepackage{collas2022_conference,times}

\usepackage{amsmath,amsfonts,bm}

\def\eqref#1{equation~\ref{#1}}

\def\1{\bm{1}}

\DeclareMathAlphabet{\mathsfit}{\encodingdefault}{\sfdefault}{m}{sl}
\SetMathAlphabet{\mathsfit}{bold}{\encodingdefault}{\sfdefault}{bx}{n}

\newcommand{\SC}[0]{\textit{Starcraft~2}}
\newcommand{\paper}[0]{Model-Free Generative Replay for Lifelong Reinforcement Learning: Application to Starcraft-2}

\usepackage{hyperref}
\usepackage{wrapfig}
\hypersetup{
    colorlinks=true,
    linkcolor=red,
    filecolor=magenta,      
    urlcolor=blue,
    citecolor=purple,
    pdftitle={\paper},
    pdfpagemode=FullScreen,
    }
\usepackage{graphicx}
\usepackage{algorithm}
\usepackage{algorithmic}

\usepackage[compact]{titlesec}  
\titlespacing{\section}{0pt}{0pt}{0pt}
\titlespacing{\subsection}{0pt}{0pt}{0pt}

\title{\paper} 

\author{Zachary Daniels, Aswin Raghavan, Jesse Hostetler, Abrar Rahman, Indranil Sur \\ \textbf{Michael Piacentino}, \textbf{Ajay Divakaran}\\
SRI International, Princeton, NJ, USA\\
\texttt{zachary.daniel@sri.com}, \texttt{aswin.raghavan@sri.com}, \texttt{jesse.hostetler@sri.com}\\ \texttt{abrar.rahman@sri.com}, \texttt{indranil.sur@sri.com} \\ \texttt{michael.piacentino@sri.com}, \texttt{ajay.divakaran@sri.com}}

\collasfinalcopy 

\begin{document}
\maketitle
\begin{abstract}
One approach to meet the challenges of deep lifelong reinforcement learning (LRL) is careful management of the agent's learning experiences, in order to learn (without forgetting) and build internal meta-models (of the tasks, environments, agents, and world).
Generative replay (GR) is a biologically-inspired replay mechanism that augments learning experiences with self-labelled examples drawn from an internal generative model that is updated over time. 
We present a version of GR for LRL that satisfies two desiderata: (a) Introspective density modelling of the latent representations of policies learned using deep RL, and (b) Model-free end-to-end learning. 
In this paper, we study three deep learning architectures for model-free GR, starting from a na\"ive GR and adding ingredients to achieve (a) and (b). 
We evaluate our proposed algorithms on three different scenarios comprising tasks from the \SC{} and Minigrid domains.
We report several key findings showing the impact of the design choices on quantitative metrics that include transfer learning, generalization to unseen tasks, fast adaptation after task change, performance comparable to a task expert, and minimizing catastrophic forgetting.  
We observe that our GR prevents drift in the features-to-action mapping from the latent vector space of a deep RL agent. 
We also show improvements in established lifelong learning metrics. 
We find that a small random replay buffer significantly increases the stability of training when combined with the experience replay buffer and the generated replay buffer.
Overall, we find that ``hidden replay'' (a well-known architecture for class-incremental classification) is the most promising approach that pushes the state-of-the-art in GR for LRL, and observe that the architecture of the sleep model might be more important for improving performance than the types of replay used. Our experiments required only 6\% of training samples to achieve 80-90\% of expert performance in most \SC{} scenarios.

\end{abstract}
\section{Introduction}


Lifelong Reinforcement Learning (LRL) involves training an agent to maximize its cumulative performance on a stream of changing \textit{tasks} over a long lifetime. LRL agents must balance \textit{plasticity vs. stability}: learning the current task while maintaining performance on previous tasks. Deep neural networks are especially prone to instability, often exhibiting \textit{catastrophic forgetting} of previous tasks when trained on multiple tasks that are presented sequentially  \citep{kirkpatrick_overcoming_2016}. One approach to meet the challenges of deep LRL is by careful managing the agent's learning experiences, in order to learn (without forgetting) and build internal meta-models (of the tasks, environments, agents, and world). One strategy for managing experiences is to recall data from previous tasks and mix it with data from the current task when training, essentially transforming sequential LRL into batch multi-task reinforcement learning (RL) \citep{brunskill2013sample}. The technique of \textit{experience replay} of past data is common in deep RL \citep[e.g.][]{mnih2013playing} and has been studied in the lifelong learning literature \citep{isele2018selective,rolnick2019experience,hayes2021replay}. 
However, storing sufficient examples from all previous tasks may be unreasonable in systems with limited resources or long lifetimes.
Rather than store an ever-growing list of experiences, one can instead employ \textit{generative replay} (GR) \citep{shin2017continual,van_de_ven_generative_2018}, a biologically-inspired replay mechanism that augments learning experiences with self-labelled examples drawn from an internal generative model trained to approximate the distribution of past experiences. 

In LRL (see \citet{khetarpal2020towards} for a survey), each task is a Markov decision process (MDP), and the agent must learn a policy for interacting with its environment to maximize its lifetime performance (e.g., average reward over all tasks). 
While GR has been shown to be effective in lifelong supervised learning \citep{van_de_ven_generative_2018}, it has been less explored in LRL. 
The LRL setting is complicated by the fact that the agent's experiences are observation-action \emph{trajectories} whose distribution depends on the environment dynamics and the policy being executed. 
While prior work with world models \citep{ketz2019continual} and pseudo-rehearsal \citep{atkinson2021pseudo} focused on generative replay of (partial) trajectories, this requires modeling the dynamics, which is difficult in complex environments. 

In this paper, we present agent architectures for LRL that use GR for memory consolidation and 
satisfy two desiderata: (a) Introspective density modelling of the latent representations of policies learned using deep RL, and (b) Model-free end-to-end learning. 
The first property avoids the challenges of density modelling of complex high-dimensional perceptual inputs, whereas policy learning using deep RL works well with such perceptual inputs. 
The second property avoids the challenges of learning temporal dynamics and reward functions from few learning experiences with sparse rewards. 
Our contributions extend powerful \emph{model-free} deep RL and do not learn the dynamics explicitly.

Our first contribution is a model-free approach to generative replay (Section \ref{sec:model-free-generative-replay}) in lifelong RL. We empirically demonstrate that replay of independently sampled observation-action pairs is sufficient to mitigate forgetting. This is noteworthy because observation-action pairs are simpler to generate than trajectories, and because replay of observation-action pairs is directly analogous to replay of input-label pairs in supervised learning, thus unifying generative replay methods for lifelong learning.

Our second contribution is a mechanism for memory consolidation via \emph{wake-sleep} cycles. During \emph{wake phases}, our agent improves a policy for the current task(s) using an off-the-shelf RL algorithm while storing samples of experiences in a buffer. Periodically, the agent enters a \emph{sleep phase}, during which the generative memory is trained to model the agent's new experiences, and the procedural knowledge embodied in the policy learned during the wake phase is consolidated into the agent's ``skill set'' via policy distillation \citep{rusu2015policy} with GR. In the next wake phase, a new exploration policy is seeded with knowledge from previously-learned skills, and the cycle repeats. 
We investigate three different replay architectures (Section \ref{sec:model-free-generative-replay}) within the wake-sleep framework, starting from a na\"ive generative and adding ingredients to achieve the aforementioned introspective density modelling of the latent representations and model-free end-to-end learning.

We evaluate our algorithm and architectural contributions through experiments (Section \ref{sec:experiments}) on three different scenarios comprising tasks from the \SC{} (SC-2) and Minigrid domains. We report several key findings showing the impact of the design choices on quantitative metrics that include transfer learning, generalization to unseen tasks, fast adaptation after task change, performance compared to  task experts, and catastrophic forgetting.   
We observe that our GR prevents drift in the features-to-action mapping of a deep RL agent (that learns the features from observations). We find that \emph{hidden replay} (replay of such features) outperforms other replay mechanisms, which agrees with past results in lifelong supervised learning \citep{van2020brain}. Since deep RL in SC-2 is computationally demanding, we release a dataset of trajectories of trained single-task expert policies for future research \footnote{\url{https://github.com/sri-l2m/l2m\_data}}.

\section{Background}
\subsection{Lifelong Reinforcement Learning (LRL) Setup}


A lifelong RL \emph{syllabus} is a sequence of \emph{tasks} that the learning agent experiences one at a time for a fixed number of interactions.
Each task is assumed to be formulated as a Partially Observable Markov Decision Processes (POMDPs) \citep{sutton2018reinforcement}, where 
each POMDP is defined by a tuple $(\mathcal{S}, \mathcal{A}, P(s'|s,a), R(s, a), \mathcal{O}, Z(o|s))$ consisting of set of states $\mathcal{S}$, a set of actions $\mathcal{A}$, a transition probability function between states $P(s'|s,a)$, a reward function $R(s, a)$, a set of observations $\mathcal{O}$, and conditional observation probability function $Z(o|s)$.

We assume that all tasks have distinct observation spaces of the same dimensionality, share a common action space (e.g., point-and-click, issue movement or attack commands), but differ in the transition and reward dynamics.
For example, this setup can capture the use case of an embodied agent using the same set of sensors and actuators throughout their lifetime but with different reward functions or objectives. 

In single-task RL, the objective is to learn a policy that maps the history of observations to actions, $\pi: \mathcal{O}^h \rightarrow \mathcal{A}$, such that the policy maximizes the expected discounted future reward, given by the value function $V^{\pi}(s) = E_{S_t \sim P, A_t \sim \pi}[\sum_{t = 1}^{\infty} \gamma^{t-1} R(S_t, A_t)]$. We will refer to the task-optimal policy as a \emph{single task expert} (STE).

Lifelong Learning involves training an agent to continually learn from a stream of tasks over long lifetimes. 
As the agent encounters new tasks, it must accumulate and leverage knowledge to learn novel tasks faster (positive forward transfer), while maintaining performance on previous tasks (minimizing catastrophic forgetting or negative backwards transfer).
In this paper, we consider the task of lifelong reinforcement learning (LRL), which has been less explored than class-incremental classification \citet{hayes2021replay} in the literature.


The objective in LRL is to learn a policy that maximizes average cumulative reward over all POMDPs in a task set while learning from limited interactions with a sequence of POMDPs drawn from the task set.
This is challenging because the learning problem is non-stationary and the learner needs to accommodate an increasing number of tasks in a model with fixed capacity.
The task changes after an unknown number of interactions with the environment. 
The agent has no control over the choice of the next task. 
In our setting the agent has no knowledge of the identity of the current task, or the transition points between tasks.

\subsection{Generative Replay (GR)}
Our work is focused on generative replay (GR), which can be broadly described as a form of data augmentation where real-world experiences are used to continually train a generative model over the input space of the problem. Subsequently, data sampled from this generator can be pseudo-labeled or self-labeled to generate paired training examples that may reflect previously trained tasks. 
Different types of replay were discussed by \citet{hayes2021replay} along with its role in the brain and in current deep learning implementations. Most of the prior work on generative replay has focused on the problem of class-incremental classification \citet{van_de_ven_three_2019,shin2017continual}, or class-incremental generative modelling \citet{cong2020gan} as a subroutine thereof, but there has been significantly less exploration of GR in LRL.

We use a Variational Auto-Encoder (VAE) \citep{kingma_auto-encoding_2014} as the generative model. The VAE consists of an encoder $q_\phi(z|o)$ mapping observations to a vector-valued latent space $z$, and a decoder $p_\theta(o|z)$ that is trained to reconstruct the original input. We use the Gaussian prior $z \sim N(0, \mathbf{I})$ with 0 mean and unit variance $\mathbf{I}$. To train the VAE, we minimize the standard loss $\mathcal{L}_\text{VAE}$, 
\begin{equation}
\label{loss:vae}
\min E_{q_{\phi}(z|o)}[\ln p_{\theta}(o|z)] - D_{KL}[q_{\phi}(z|o)||p(z)]
\end{equation}
World models \citep{ketz2019continual} use model-based generative replay for LRL i.e., the replay generates rollouts or entire trajectories \citep[e.g.][]{ketz_using_2019}. Model-based replay works well in many domains, but it is challenging for domains such as SC-2 due to the complex dynamics and high dimensional observation and action spaces.
The most relevant prior work is that of \citeauthor{atkinson2021pseudo} \citep{atkinson2021pseudo} who explored model-free generative replay for Deep Q-Learning (DQN). 
However, \citet{atkinson2021pseudo} uses GANs in the raw perceptual space of the agent (aka veridical replay \citet{hayes2021replay}). Others have explored sophisticated strategies for prioritized experience replay in DQN  \citep[e.g.][]{isele2018selective, riemer2019scalable,wang2021acae,caccia2020online,zhang2017deeper,schaul2015prioritized} in LRL, but these are not generative in nature. 
We show that na\"ively migrating the generative modelling (similar to \citeauthor{atkinson2021pseudo}) to the feature space of the policy causes instabilities and severe catastrophic forgetting. 
To the best of our knowledge, this is the first work on model-free GR for LRL in the feature space of generic deep RL agents. 

\section{Hidden Replay for Lifelong RL}
\label{sec:model-free-generative-replay}
This section introduces our core contribution of wake-sleep training combined with model-free GR of latent representations. We also explore two other GR architectures along the way. 
 The fundamental tradeoff in lifelong learning is between \emph{plasticity}, i.e.\!, learning the current task, and \emph{stability}, i.e.\!, remembering past tasks.
We address this tradeoff in lifelong RL by employing an approach that alternates between two distinct phases \citep{raghavan2020lifelong}.
In the wake phase, a \emph{plastic} wake policy $\pi_w$ is optimized for the current task by interaction with the environment and using any off-the-shelf RL method.
Transition tuples are collected during training and stored in a buffer (see Alg. \ref{alg:wakesleep}); each tuple contains $(o, r, a)$, the observation $o$, reward for the previous action $r$, and the policy output $a$ (e.g., the policy logits or one-hot encoded action).
The sleep policy $\pi_s$ provides ``advice'' to explore the current task using the consolidated policy learned from all previous tasks.
Wake phase actions are selected according to a mixture of $\pi_w$ and $\pi_s$, the probability of choosing an action from $\pi_s$ decaying to 0 over time.
We use an off-policy RL algorithm such as VTrace \citep{espeholt2018impala} to accommodate this off-policy action selection in the optimization of $\pi_w$.

\begin{figure}[t]
\centering
\includegraphics[width=0.92\textwidth]{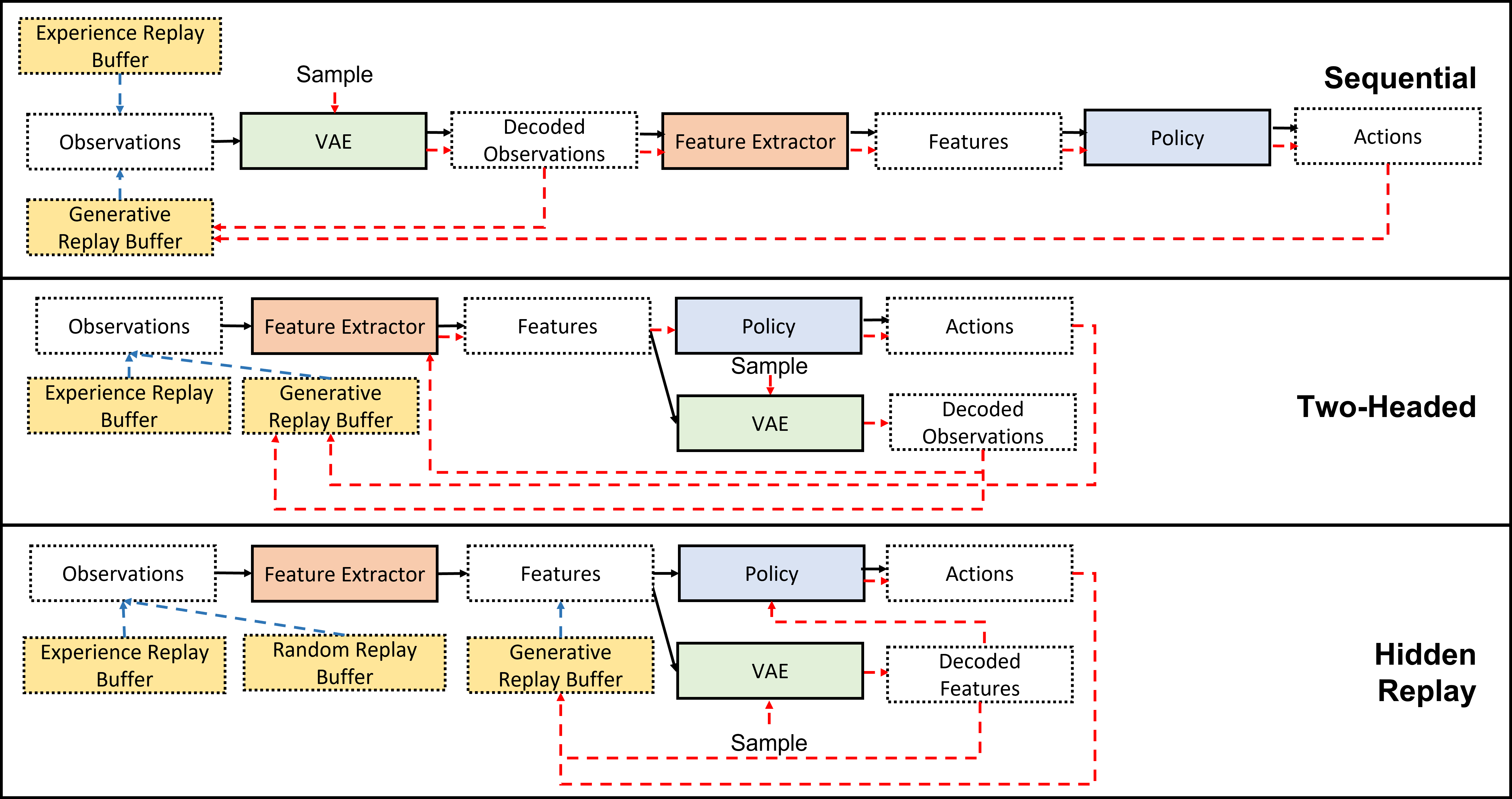}
\caption{We investigate three replay architectures for model-free generative replay (sleep phase). The red dashed arrows denotes sampling steps starting from the VAE to generate labelled actions. The blue dashed arrows show how the various replay buffers are sampled and  fed into the sleep model during training. The black arrows denote the forward pass. \textbf{Top:} Sequential Replay Architecture; \textbf{Middle:} Two-Headed Replay Architecture; \textbf{Bottom:} Hidden Replay Architecture.}
\label{fig:all_replay_archs}
\end{figure}

In the sleep phase, the \emph{stable} sleep policy $\pi_s$ is optimized with the objective of incorporating new knowledge (the action selection in the wake buffer) while retaining current knowledge. 
Similar to generative replay in supervised learning, augmented dataset(s) are created by combining wake transitions with tuples from the generative model $g_s$ and pseudo-labelled by the sleep policy. 
This form of replay is model-free because it does not learn the POMDP dynamics and reward and does not generate and replay temporal trajectories.  
The sleep policy and the generative model are  jointly trained.
The sleep policy is trained to minimize $\mathcal{L}_{xent}$, a standard cross-entropy loss between the outputs (logits) of $\pi_s$, treating the replay augmented dataset as the ground truth. 
The generative model is trained using the combined dataset and the standard VAE loss (Eq. \ref{loss:vae}). 
 The description so far is easily applied to veridical replay, i.e.\!, when generative modelling and replay occurs in the original observation space of the MDP, but this is undesirable as discussed before. The design choice of whether the input or hidden space is replayed distinguishes the model implementations (Alg. \ref{alg:wakesleep} vs Alg. \ref{alg:hidden_replay}, and models in Fig. \ref{fig:all_replay_archs}) from each other. 

   \begin{minipage}{0.49\textwidth}
\begin{algorithm}[H]
   \caption{Wake-Sleep Training in the Sequential and Two-Headed models.} 
   \label{alg:wakesleep}
\begin{algorithmic}[1]
   \STATE {\bfseries Iterates:} Generator $g_s^t$, sleep policy $\pi_{s}^t$, wake policy $\pi_{w}^t$, wake buffer $b^t_w$
	   \FOR{$t=1,2,\ldots$}
	   \FOR[Wake Phase]{$K$ times}
	        \STATE{Sample observation $o$ and reward $r$ from current task}
	       \STATE Sample $a \sim \pi_s^t(o)$ w.p. $p_{\text{advice}}$, else $a \sim \pi_w^t(o)$
	       \STATE{Add transition $(o, r, a)$ to FIFO wake buffer $b_w^t$}
	       \STATE{Update wake policy $\pi^t_w$ on current task reward using $b^t_w$}
		   \STATE{Decay $p_\text{advice}$}
	   \ENDFOR\\ 
	   \FOR[Sleep Phase]{$N$ iterations} 
		   \STATE{Sample batch $B$ from $b_w^t$ (c.f.\! Line 6).} 
			 \STATE Sample batches $O_s^t \sim g_s^t$
			 \STATE Pseudo-label $A_s^t = \pi_s^t(O_s^t)$
			 \STATE (GR Buffer) $D = B \cup (O_s^t, A_s^t)$
			 \STATE Minimize $\mathcal{L}_{VAE}(g_s^{t+1}) + \alpha\mathcal{L}_{xent}(g_s^{t+1}, \pi_s^{t+1})$ on $D$
		\ENDFOR\\
	 \ENDFOR
\end{algorithmic}
\end{algorithm}
\end{minipage}
\hfill
\begin{minipage}{0.49\textwidth}
    \begin{algorithm}[H]
       \caption{Wake-Sleep Training in the  Hidden Replay model.} 
       \label{alg:hidden_replay}
    \begin{algorithmic}[1]
       \STATE {\bfseries Iterates:} Generator $g_s^t$, sleep policy $\pi_{s}^t$, wake policy $\pi_{w}^t$, wake buffer $b^t_w$, random replay buffer $b^t_r$
    	   \FOR{$t=1,2,\ldots$}
    	   \FOR[Wake Phase]{$K$ times}
       	        \STATE{Sample observation $o$ and reward $r$ from current task}
    	       \STATE Sample $a \sim \pi_s^t(o)$ w.p. $p_{\text{advice}}$, else $a \sim \pi_w^t(o)$
    	       \STATE{Add transition $(o, r, a)$ to wake buffer $b^t_w$}
    	       \STATE{Update wake policy $\pi^t_w$ on current task reward using $b^t_w$}
    		   \STATE{Decay $p_\text{advice}$}
    	   \ENDFOR\\ 
    	   \FOR[Sleep Phase]{$N$ iterations}
    		   \STATE{If ER: Batch $B \sim b^t_w$ (c.f.\! Line 6).} 
    			 \STATE{If GR: Feature batches $H^t_s \sim g_s^t$}
    			 \STATE{If GR: Pseudo-label $A^t_s \sim \pi_s^t(H_s^t)$}
    			\STATE{If RaR: $b^t_r = b^t_r \bigcup b^t_w.sample(K)$}
    			\STATE{If RaR: $D_r \sim b^t_r$}
    			 \STATE Minimize $\mathcal{L}_{VAE}(g_s^{t+1}) + \alpha\mathcal{L}_{xent}(g_s^{t+1}, \pi_s^{t+1})$ on $(H^t_s, A^t_s)$ and end-to-end minimize (including feature extractor) the same loss on observation-action pairs in $B \bigcup D_r$
    		\ENDFOR\\
    	 \ENDFOR
    \end{algorithmic}
    \end{algorithm}
    \end{minipage}
    \hfill

We compare three replay architectures that we denote Sequential, Two-Headed, and Hidden Replay (shown in Fig. \ref{fig:all_replay_archs}). Each architecture is a specific arrangement of the following three components: a feature extractor, a VAE, and a policy network. The feature extractor maps raw image-like observations to feature vectors using a convolutional neural network (CNN) or a multi-layered perceptron. The policy network is dependent on the action space, e.g.,  in our experiments we utilize a convolutional architecture to output spatial actions.

In the \hypertarget{text:sequential}{\textbf{Sequential}} architecture (Fig.~\ref{fig:all_replay_archs}, Top), the VAE is upstream of the policy network, and the sleep policy $\pi_s$ operates on the VAE reconstruction rather than the true observation. This architecture is similar to the one proposed in \citet{atkinson2021pseudo}.
During sleep (following the red dashes in Fig.~\ref{fig:all_replay_archs}, Top), sampled latent vectors are decoded to observations that are then labelled by the policy. We hypothesize that the sequential architecture tends to map observations for which the policy chooses similar actions to generic reconstructions. We found that this method produces blurry reconstructions for image-like observations. 

The \hypertarget{text:twoheaded}{\textbf{Two-Headed}} architecture (Fig.~\ref{fig:all_replay_archs}, Middle) decouples the decoder/generator from the policy network. Each observation (real or generated) is passed through the feature extractor. The features are then separately passed through the VAE and the policy network. Our experiments showed improved visual reconstructions (see Appendix \ref{sec:reconstructions_app}) due to separation of the influence of the policy on the VAE decoder. However, sampling in this architecture requires two forward passes as shown in the figure (once to decode an observation and once to label it). 

Alg. \ref{alg:wakesleep} shows the data flow that is common between these two architectures that perform veridical replay, i.e.\!, the VAE posterior is a density over raw and complex perceptual space. The training is wasteful because the wake policy $\pi_w$ has already learned an effective feature extractor, and latent-to-action partition that works well for the current task (solved in wake phase). The sleep phase in veridical replay does not utilize these learned features and tries to re-learn them, whereas deep RL (as in wake phase) is effective at learning representations to map raw percepts to actions.   


\setlength{\columnsep}{5pt}%
\begin{wrapfigure}{R}{0.35\textwidth}
\centering
\includegraphics[width=0.35\textwidth, keepaspectratio]{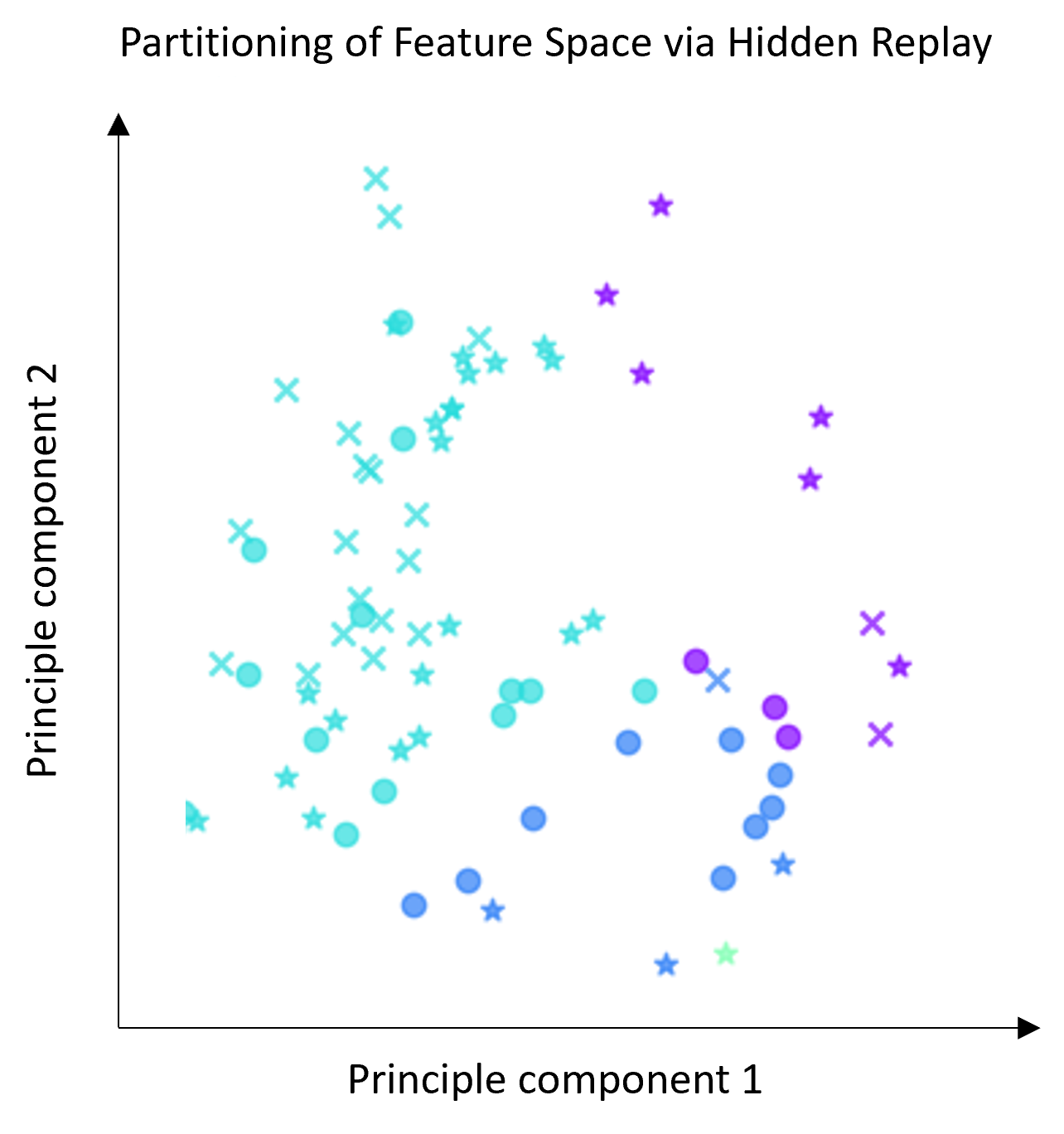}
\vspace{-0.75cm}
\caption{Hidden replay prevents drift in the features-to-action
mapping from the latent vector space of a deep RL agent (Minigrid tasks). Colors represent actions; $\times$s: drawn from the random replay buffer, $\bullet$: drawn from wake buffer; $\star$: sampled via generative replay.}
\vspace{-0.75cm}
\label{fig:hidden_replay_partitioning}
\end{wrapfigure}

The final architecture we consider is inspired by the success of \hypertarget{text:hidden}{\textbf{Hidden Replay}} in class incremental learning \citep{van2020brain} and its connection to biology. This architecture (Fig.~\ref{fig:all_replay_archs}, Bottom) generates feature vectors directly rather than observations. Sampling features instead of image-like observations benefits from dimensionality reduction. Reconstructing or generating complex image observations is hard and requires increasingly complex generative models \citep{van2017neural,van2016conditional,razavi2019generating}. In RL, especially when working with ``symbolic'' (semantic) (as opposed to ``perceptual'' (RGB)) observations, changing a small detail of the observation can change the optimal action drastically. Observations might be better separated in feature space, and thus, the sleep policy could be less sensitive to an imperfect decoder. In addition, in contrast to the two-headed architecture, the feature extractor does not need to be run when sampling. 

For example, when faced with new tasks that require new actions to be taken, hidden replay can learn to add a new action partition to the feature space while rearranging the existing feature space in a manner that preserves partitioning needed to solve previous tasks. Empirically, we observe a clean partition of the feature space according to the optimal action. Even though samples and reconstructions are not perfect, they fall within the same action partition (Fig.~\ref{fig:hidden_replay_partitioning}) as the ground truth. These representations may allow for knowledge transfer between tasks because tasks requiring similar actions can potentially efficiently be mapped into the existing feature space, but validating this requires further studies. 

Hidden replay is not without limitations because GR samples are in terms of features without the corresponding observations. This V-shape in the training graph (blue lines in Fig. \ref{fig:all_replay_archs}, Bottom) can cause the feature extractor to drift over time or collapse to a mode (e.g., map all observations to the same feature vector). 
As new tasks are encountered, this would cause sampled features 
to not align with the features mapped by the policy and VAE components. 

To limit feature drift, we store a small (typically $K < 1\%$ of the experience replay buffer) set of raw observations paired with actions in a separate buffer. We denote this RaR (for  ``random replay'' buffer). These are accumulated over the lifetime of the agent from all wake replay buffers $b^t_w$ (although a more sophisticated exemplar selection \citep{rebuffi2017icarl,mi2020ader} is also possible). We observed that replaying a few random examples results in less feature drift between task changes (See Appendix ~\ref{sec:justification_of_exemplars} for an ablation). In our experiments, we found that saving 96 random samples per sleep was sufficient to alleviate feature drift in the \SC{} domain and saving 256 random samples per sleep was sufficient for the Minigrid domain. The complete data flow for hidden replay for LRL is shown in Alg. \ref{alg:hidden_replay}, where the design choice of combinations of ER, GR, and RaR buffers are ablated in our experimental ablations.

\textbf{Remark}: When applied to single task RL, our wake-sleep cycle resembles policy iteration interleaved with policy distillation steps, where each iteration $t$ produces an improved $\pi_w^t$ (nudge towards STE) that is then distilled into $\pi_s^t$. Wake experiences are gathered from the most recent sleep policy (``advice''), including from multiple sub-optimal policies similar in spirit to dataset aggregation in DAGGER \citep{ross2011reduction}. Similar to DAGGER, the consolidated policy is executed to sample next trajectories. Our supplementary experiments (Appendices \ref{sec:wake_additional_results} and \ref{sec:ste_additional_results})  show stable convergence of the wake-sleep cycle to optimal policies in single task training (convergence to STE). 
\section{Experiments}
\label{sec:experiments}
\subsection{Evaluation scenarios}
\label{sec:scenarios}
A lifelong learning \textit{scenario} describes the order of task presentation and repetition of tasks. A concrete instantiation of a scenario with tasks (or POMDPs) is called a \textit{syllabus}. Each scenario is used to evaluate the learning agent over all syllabi and average metrics. The three  scenarios we consider are depicted in Figure~\ref{fig:scenarios}.
 \textbf{Pairwise:} Two tasks are presented once each. This scenario evaluates knowledge transfer between two tasks. \textbf{Alternating:} Two tasks are presented with three repetitions of each pair. This scenario evaluates knowledge retention over longer lifetimes. \textbf{Condensed:} A random permutation of all tasks is presented. 
 This is a useful setting for understanding how an agent performs across different task orderings.
 
Given a syllabus, an experiment run alternates between \emph{evaluation blocks} (EBs) and \emph{learning blocks} (LBs) \citep[see][]{new2022lifelong}. 
We say that a task is \emph{seen} wrt an EB if it has appeared in any LB preceeding it, otherwise it is \emph{unseen}.
During each EB, the average accumulated reward of the agent is evaluated on all tasks in the syllabus (including unseen tasks); during each LB, the agent learns on a single task for a fixed number of interactions. 
The STE for each task serves as a baseline and measures the relative performance of the learner wrt asymptotic optimal.

\subsection{Metrics for Lifelong RL}
We run controlled and quantitative experiments and compare to multiple baselines along three sets of metrics. 
First, we compare learning behavior by examining the accumulated rewards captured by periodic evaluations of the agent, i.e., we look at the trend of EB rewards averaged over EBs (and tasks). 
Second, we aim to understand how the rewards achieved by an agent compares to 
a single task expert (STE). 
This forms the basis for our second set of metrics: \emph{relative reward (RR)}.
Note that these metrics also focus on 
EBs and do not consider  
LBs. Finally, we compare algorithms using the lifelong learning metrics defined by \citep{new2022lifelong}, which take into account both LBs and EBs, and aim to frame metrics in a way that are agnostic to the domain.

We begin by describing the RR metrics. 
For each task, we train (offline) the STE 
using an off-the-shelf deep RL algorithm. 
RR metrics measure how well the agent has learned the seen tasks, how well the agent performs over all tasks in the syllabus, and how well the agent generalizes to unseen tasks. 
In general, RR is computed as
\begin{align}
RR(\pi,T,S) &= \frac{1}{|T|}\sum_{t \in T}\frac{1}{|S(t)|}\sum_{s \in S(t)} \frac{r_\pi(s)}{r_{\pi^*_s}(s)} 
\label{eq:rr}
\end{align}
where $\pi$ is the policy of the learner, $T$ is a set EBs, $S(t)$ is the set of tasks at EB $t$, $\pi^*_s$ is the STE for  task $s$, and $r_{\pi}(s)$ is the accumulated reward (return) achieved on task $s$ from using policy $\pi$. Different metrics are defined by setting $S, T$ to different EBs and tasks. 

$\mathbf{RR_\Omega}$: $T$ only includes the final EB, and $S(t)$ is the set of  seen tasks in the syllabus. $RR_\Omega$ measures how well the agent has learned the seen tasks in the syllabus ``at the end of the day''. To measure catastrophic forgetting, we denote $\mathbf{RR_\sigma}$ where 
$T$ includes all EBs, and $S(t)$ is the set of seen tasks in the syllabus encountered in previous LBs up to and including EB $t$. $\mathbf{RR_\upsilon}$ is complementary in that it measures how well the agent generalizes to unseen tasks, $T$ includes all EBs, and $S(t)$ is the set of unseen tasks not encountered in previous LBs up to and including EB $t$. To measure plasticity, we denote $\mathbf{RR_\alpha}$ where $T$ is the set of all EBs, and $S(t)$ is only the task learned in LB $t$ preceeding EB $t$.
Note that in all cases, higher more-positive values are better for all metrics as all our STEs achieve positive return.

As mentioned, we employ established lifelong learning metrics defined by \citet{new2022lifelong}, which capture catastrophic forgetting, transfer between all pairs of tasks, and speed of learning. In all cases, more-positive values are better for all metrics. 
\textbf{Forward Transfer Ratio (FTR)} is the ratio of EB performance on an unseen task. \textbf{Backward Transfer Ratio (BTR)} is the ratio of EB performance on a seen task. A value greater than one indicates positive transfer. \textbf{Relative Performance (RP)} is the ratio of the Area Under the Learning Curve (AUC) between the lifelong learner and STE learning curve. A value greater than one indicates either faster learning by the lifelong learner and/or superior asymptotic performance. \textbf{Performance Maintenance (PM)} is the average difference between task performance in any EB that comes after the LB of a seen task. 
A value less than $0$ indicates forgetting. Refer to \citep{new2022lifelong} for more details about the metrics and the precise computation.

\begin{figure}[t]
\centering
\includegraphics[width=\textwidth]{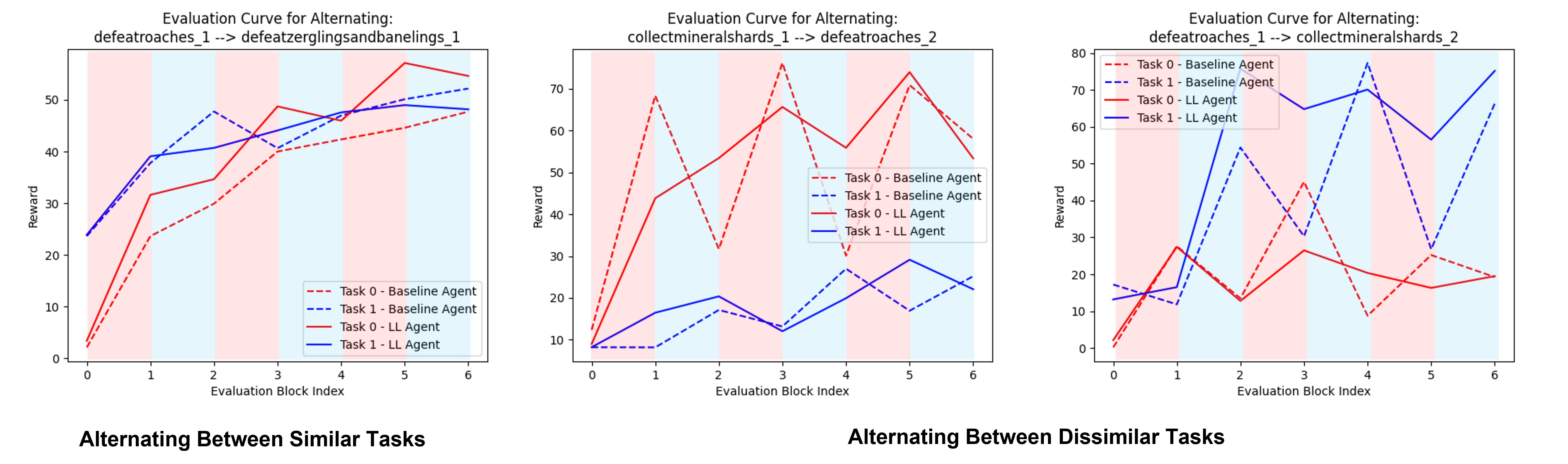}
\caption{Illustrative examples showing EB rewards of baseline and hidden replay in alternating tasks scenario. Column color denotes which task is trained in LBs between EBs. Left: Example syllabus of two similar tasks; the lifelong learner (LL) performs similarly to the baseline.
Right: Examples of two dissimilar tasks where the baseline exhibits catastrophic forgetting, our hidden replay does not.}
\label{fig:dissimilar_alternating}
\vspace{-0.75cm}
\end{figure}

\subsection{Domains} 

Our experiments use two RL domains: \SC{} minigames \citep{vinyals2017starcraft} and Minigrid \citep{gym_minigrid}. We evaluate on existing RL tasks (POMDPs) in these domains. 

\textbf{Starcraft 2}: 
We selected 3 minigames namely \emph{DefeatRoaches},  \emph{DefeatZerglingsAndBanelings}, and \emph{CollectMineralShards}. We created an additional task in each case as described in Appendix ~\ref{sec:sc2_task_descriptions}. We used PySC2 \citep{vinyals2017starcraft} to interface with SC-2 game. 
The agent receives positive rewards for collecting resources and defeating enemies, and negative rewards for losing friendly units. 
As observations, we used the unit type, selection status, and unit density two-dimensional inputs. 
We follow the action space and convolutional policy network architecture defined in \citet{vinyals2017starcraft}. 
Single task experts were trained to convergence ($\sim$ 30M environment steps using the VTrace algorithm \citep{espeholt2018impala}). We consider the pairwise, alternating, and condensed scenarios where each LB consists of 2 million environment steps (about 6\% of the STE samples), and each EB consists of 30 episodes per task. Complete details can be found in the Appendix \ref{sec:imp_sc2}.

\textbf{Minigrid}: 
Due to the heavy computational load of experimentation in SC-2, we also show results in the Minigrid domain \citep{gym_minigrid}. We use the TELLA framework \citep{lifelong-learning-systems} to orchestrate the experiments ($\sim$100 repetitions per scenario). Tasks involve an agent that must navigate to a goal while interacting with various entities in the environment. We consider 5 tasks --- SimpleCrossing, DistShift, a custom Fetch, a custom Unlock, and DoorKey --- each with two variants (differing in the size of the grid and/or number of objects) for a total of 10 POMDPs. Descriptions of these tasks can be found in \citet{gym_minigrid}. An observation is a $7\times7$ top-down view of the grid directly in front of the agent, with 3 channels containing information about the object type, object color, and object state in each tile. Rewards are sparse: $-1$ for running into obstacles and lava, $0$ for not reaching the goal, and a reward in $[0, 1]$ for reaching the goal proportional to the number of steps taken. There are only six actions: turn left, turn right, move forward, pick up an object, drop an object, and interact with an object. STEs are trained on each task for 1M environment steps using PPO \citep{schulman2017proximal}. We consider the pairwise and condensed scenarios where the LB lengths vary between 100k and 700k (Appendix~\ref{sec:minigrid_task_lengths}) based on examining the smallest number of steps for the STE to converge to $80\%$ of its peak performance averaged over ten runs. Each EB consists of 100 episodes per task. Further details can be found in Appendix~\ref{sec:imp_minigrid}.

\textbf{Baseline Agents}:
We compare against several non-lifelong baseline agents: 1) Random policy; 2) \emph{Baseline agent} using off-the-shelf deep RL trained on tasks in sequence without any lifelong learning. We used the VTrace algorithm \citep{espeholt2018impala} for SC-2 and PPO  \citep{schulman2017proximal} for the Minigrid experiments. 

\textbf{Ablations}: In order to compare different replay type, we perform ablations against the combination of buffers used for replay (see Fig. \ref{fig:all_replay_archs}): 1) \emph{Hidden Replay (ER)}, 2) \emph{Hidden Replay (ER+RaR)}, 3) \emph{Hidden Replay (ER+GR)}, 4) \emph{Hidden Replay (ER+RaR+GR)}. In each case, the VAE, sleep policy, and wake policy are trained, but on different augmented datasets as shown in Alg.~\ref{alg:hidden_replay} (samples are drawn from VAE only if GR is enabled).
Similarly, we perform ablations along architectures discussed in Fig.~\ref{fig:all_replay_archs}.

\textbf{Single Task Experiments}: 
We conducted experiments that demonstrated that the wake-sleep training phases successfully imitates the STE. These can be found in the Appendix \ref{sec:ste_additional_results} for SC-2 and Minigrid. 

\subsection{Evidence of Overcoming Catastrophic Forgetting}
We observed that different syllabi and scenarios can exhibit different levels of catastrophic forgetting due to varying degrees of similarity between the tasks in the syllabus. We first illustrate this issue before showing large scale results averaging over all syllabi.\footnote{We attempt to quantify pairwise similarity between tasks in Appendix \ref{sec:imp_sc2} by evaluating the forward transfer of the STE for each task to all other tasks. Quantifying task similarity remains an open problem.}
In Figure \ref{fig:dissimilar_alternating}, we illustrate examples that 
1) the LL agent and non-LL agent perform similarly when syllabi are composed of similar tasks, and both agents exhibit positive FT and BT, and 2) the non-LL agent experiences catastrophic forgetting (zig-zag pattern) of one or both tasks whereas the LL agent shows significantly less forgetting \emph{when transitioning between two dissimilar tasks}. 
We note that in the \SC{} domain, evidence of extreme catastrophic forgetting is somewhat rare, a possible weakness in the design of our task set. Nevertheless, LL agents clearly outperform non-LL on almost all the other  metrics, signifying the need to look beyond catastrophic forgetting when evaluating LL. 

\subsection{Comparison of Different Types of Replay for Sleep}
\begin{figure*}[t]
\centering
\includegraphics[width=.98\textwidth]{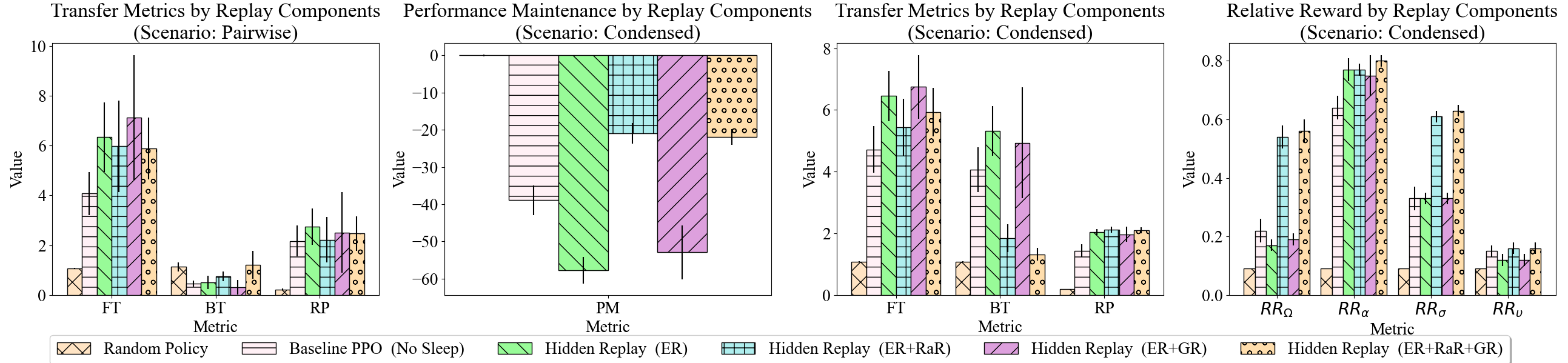}
  \caption{Evaluating different replay types on the hidden replay model for the Minigrid domain. Error bars denote 95\% confidence interval.}
  \label{fig:minigrid_replay_ablations_main}
\end{figure*}


\begin{figure*}[t]
\centering
    \includegraphics[width=0.98\textwidth]{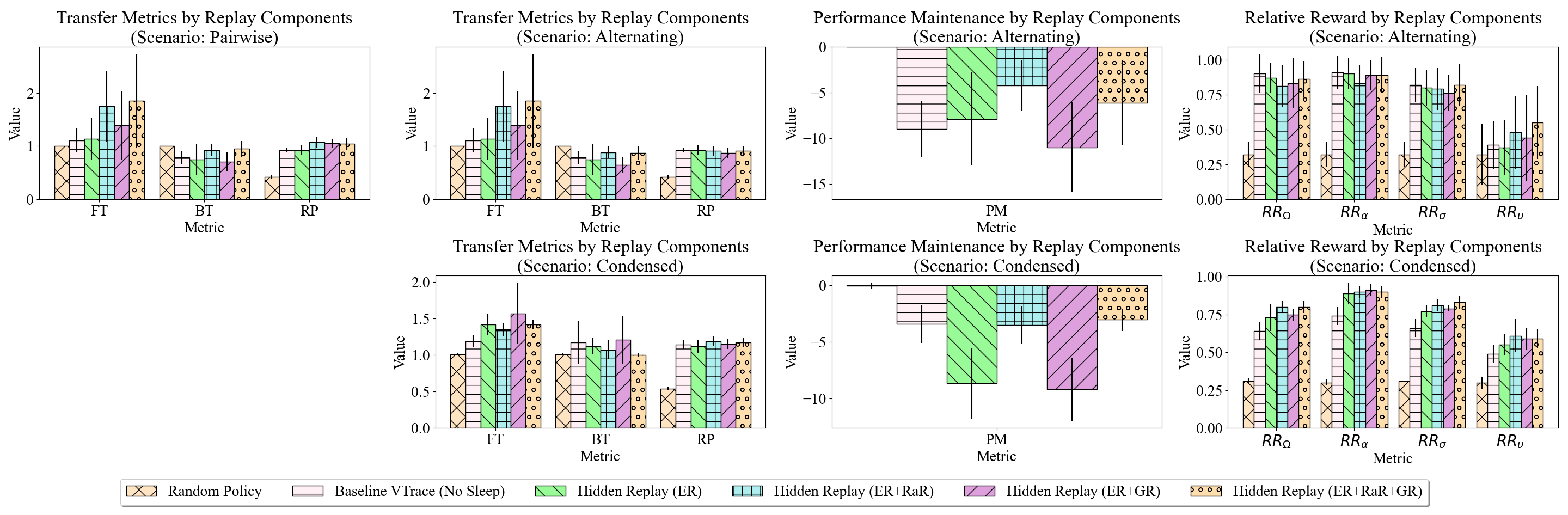}
  \caption{Evaluating different replay types on the hidden replay model for the \SC{} domain. Error bars denote 95\% confidence interval.}
  \label{fig:sc2_replay_ablations_main}
\end{figure*}

To understand the effect of different replay types on the hidden replay model: experience replay from the wake buffer (ER), generative replay (GR), and random replay (RaR). All approaches still train both the policy network and generator, even if GR is disabled. We summarize the important findings in Fig.~\ref{fig:minigrid_replay_ablations_main} and Fig.~\ref{fig:sc2_replay_ablations_main} \footnote{Further experiments can be found in Appendix \ref{app:replay_ablation}}.

\textbf{Transfer Learning and Generalization to Unseen Tasks}: 
Looking at the pairwise scenario in Minigrid and \SC{}, we see that models that use hidden replay with ER+GR, ER+RaR or ER+RaR+GR show much higher FT compared to the baseline. Hidden replay with ER+GR, ER+RaR or ER+RaR+GR show higher $RR_\upsilon$ and the effect is more pronounced for \SC{}  condensed scenario, showing generalization to unseen tasks. 

\textbf{Mitigating Catastrophic Forgetting}:
Looking at PM across all scenarios in Minigrid and \SC{} and looking at  $RR_\Omega$ of the Minigrid condensed scenario, we validate that the hidden replay model requires stabilization with RaR in order to (significantly) reduce catastrophic forgetting, compared to ER+GR alone which is typically worse than ER+RaR and ER+RaR+GR.

\textbf{Adaptation to New Tasks}: Interestingly, we find that $RR_\alpha$ for the condensed scenarios of Minigrid and \SC{}, shows that simply performing distillation during sleep phases produces 
a beneficial ``jumpstart'' effect when learning new tasks, compared to the baseline which is also able to learn the new task but requires more learning experiences. 

\textbf{Backward Transfer}: The BT metric seems to vary across the board based on the type of replay used. This is in part because when computing the BT metric from task 2 to task 1, it is very sensitive to how well the model learns on task 1. If the model learns task 1 to near optimal performance, it is very difficult to achieve $BT>1$ whereas if task 1 is not well-learned, even a weak model on a similar task 2 can potentially easily achieve $BT>1$. However, if we look at $RR_\sigma$ instead, we see a different story where the models that learn best (ER+RaR and ER+RaR+GR) often exhibit the best performance on seen tasks in opposition to the BT metric.

\textbf{General Observations}: 
Adding a small amount of random replay (ER+RaR+GR) fixes many issues with standard generative hidden replay, producing noticeably better results. We find that ER+RaR, even without GR, achieves significant improvement over the baseline agent, and the model further improves when all types of replay (ER+RaR+GR) are combined, albeit by a small amount. This suggests that RaR of a small number of samples is often powerful and 
complementary (i.e., often helpful, rarely hurtful) to GR (n.b.\! all methods train the VAE, but only GR samples from it). This leads us to suspect that the replay architecture is more important than the sampling strategy that we investigate next. 


\subsection{Comparison of Different Architectures for Sleep}

\begin{figure*}[t]
\centering
    \includegraphics[width=0.98\textwidth]{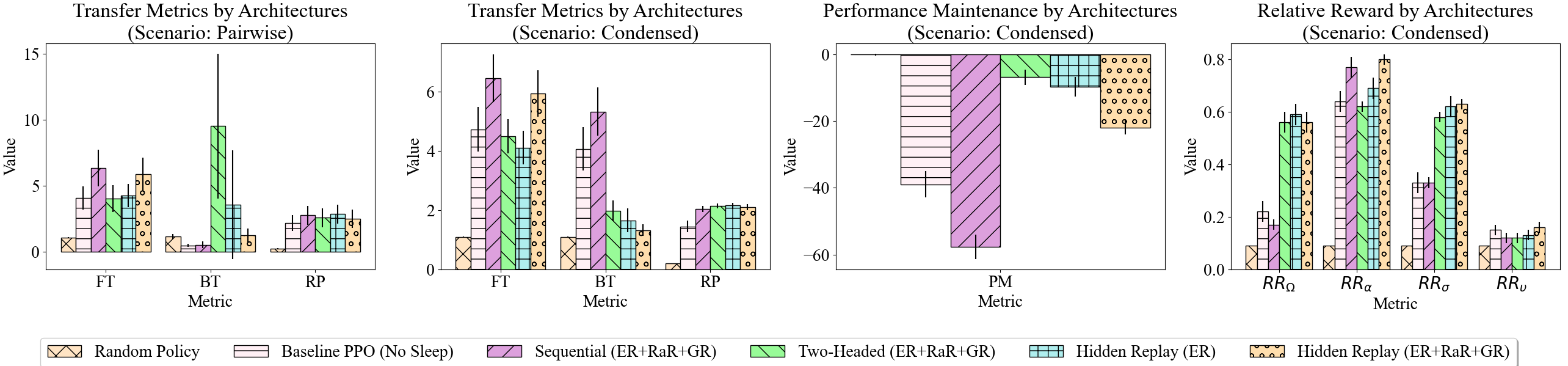}
  \caption{Evaluating the performance of different architectures for the Minigrid domain. Error bars denote 95\% confidence interval.}
  \label{fig:minigrid_ablations_archs_main}
\end{figure*}

\begin{figure*}[t]
\centering
    \includegraphics[width=0.98\textwidth]{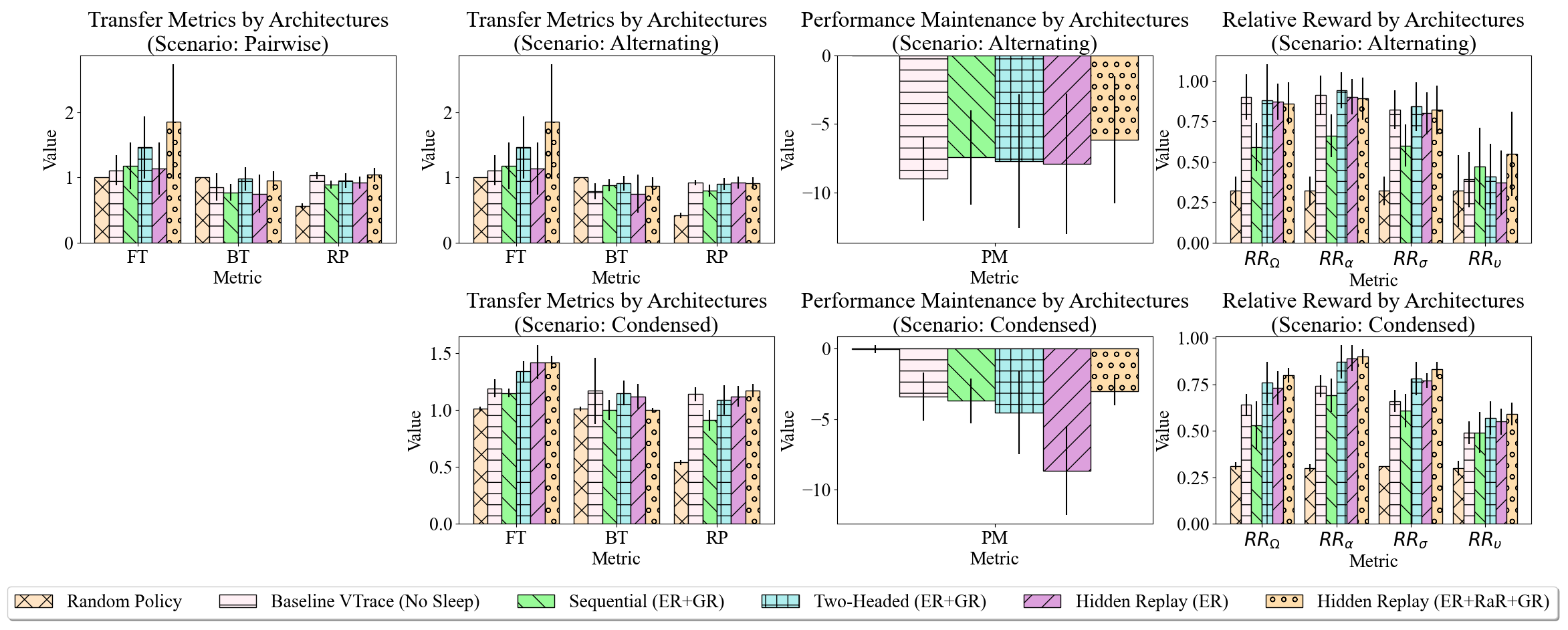}
  \caption{Evaluating the performance of different architectures for the \SC{} domain. Error bars denote 95\% confidence interval.}
  \label{fig:sc2_ablations_archs_main}
  \vspace{-0.5cm}
\end{figure*}

We aim to understand the strengths and weaknesses of the model architectures introduced in this paper (Sequential, Two-Headed, Hidden Replay) and understand how these models improve over simple baselines. 
We summarize the important findings in Fig.~\ref{fig:minigrid_ablations_archs_main} and Fig.~\ref{fig:sc2_ablations_archs_main} \footnote{Further experiments can be found in Appendix \ref{app:archs_ablation})}.

\textbf{Transfer Learning and Generalization to Unseen Tasks}: Forward transfer seems to be a strength of the hidden replay model.  Looking at FT across all domains and scenarios and looking at the $RR_\upsilon$ of the \SC{} scenarios, we see that the hidden replay model noticeably outperforms the other models.

\textbf{Avoiding Catastrophic Forgetting}:
Looking at PM and $RR_\Omega$ across all scenarios, the hidden replay model often outperforms the other models.

\textbf{Adaptation to New Tasks}: We see that the hidden replay model achieves consistently high (and often the best) RP and $RR_\alpha$ across most scenarios in both domains. This suggests that the hidden replay model is not interfering with new tasks and may be providing a helpful ``jumpstart'' effect. 

\textbf{Positive Backward Transfer}: As before, it can be difficult to interpret the BT metric. In general, the BT of the hidden replay model is on par with other models that achieve comparable terminal performance, and the hidden replay model shows strong performance in terms of $RR_\sigma$.

\textbf{Sample Efficiency}
Compared to STE, our experiments using the hidden replay model required only 6\% of training samples to achieve 80-90\% of expert performance in most \SC{} scenarios.

\textbf{General Observations}: In general, across both domains and many scenarios, the hidden replay model seems to outperform the other model architectures tested in most of the metrics of interest. This suggests that the hidden replay model is a potentially powerful model for addressing lifelong RL. We observe (in the case of the Minigrid experiments where all types of replay are used for all architectures), architecture seems to play a more significant role in improving lifelong learning performance than the type of replay. 
\section{Discussion}


Our experiments demonstrate LRL via positive FT and BT, high performance relative to an STE, reduced catastrophic forgetting, and fast adaptation to new tasks. Our work clearly advances the domain-specific state-of-the-art in \SC{} by reducing the sample complexity of deep RL to master the minigames.
Compared to STE, our experiments required only 6\% of training samples to achieve 80-90\% of expert performance. 
Existing lifelong RL implementations (e.g., \citep{mendez2020lifelong}) have been tested on benchmark problems, but proved to be challenging to scale up to the complex observation and action space of SC-2. Future work can leverage our techniques in the full SC-2 game and reduce the astronomical sample complexity of current deep RL agents in SC-2 (e.g., AlphaStar \citep{vinyals2019grandmaster}). 

While the wake-sleep mechanism and model-free GR show promising results, there is room for improvement, and the approach is easily extensible. We highlight several potential extensions. Our current model triggers sleep on a set schedule. Instead, it is possible to self-trigger sleep via recent work in lifelong changepoint detection \citep{faber2021watch}. Currently, we store full observations in the replay buffer, which due to memory constraints, limits the number of samples that can be saved. It would be prudent to intelligently compress experiences in the replay buffer \citep{riemer2019scalable, wang2021acae, caccia2020online,zhang2017deeper,schaul2015prioritized, hayes2019remind}. Past work has looked at improving replay buffers via prioritized replay \citep{schaul2015prioritized} based on detecting dead-end states. 
We further hypothesize that increasing the lifetime of an agent by avoiding dead end states is a useful bias. Our methods operate over hand-crafted representations of the observation space; the system could be extended by introducing representation learning from RGB-observations via unsupervised (STAM \citep{smith2019unsupervised,smith2019unsupervised2}) and self-supervised learning \citep{gallardo2021self}. We can further extend our models using mixtures of VAEs (e.g., the Eigentask framework \cite{raghavan2020lifelong}) to dynamically grow and shrink the model, leading towards agents that build implicit models of the tasks they experience. Finally, we used a fixed neural architecture and loss functions throughout the agent's lifetime. GR is a flexible approach that can include progress in other areas of lifelong learning, e.g., dynamically changing the underlying neural architectures \citep{schwarz2018progress}, neuromodulation \citep{brna2019uncertainty}, and regularization \citep{kolouri2019sliced}. 
Further, interesting avenues include the application of GR to multi-agent RL \citep{nekoei2021continuous} and federated learning \citep{rostami2017multi}, where communication between agents can enable better coordination between agents and collective acquisition of knowledge. Due to the inherent plastic-yet-stable nature of lifelong learners, our methods can be more robust to data drift, concept drift, or task changes, reducing system downtime and the overhead of retraining, and used to address  model obsolescence and reduce the technical debt of machine learning models \citep{sculley2015hidden, alahdab2019empirical, bogner2021characterizing}. 




\subsubsection*{Acknowledgments}
This material is based upon work supported by the Defense Advanced Research Projects Agency (DARPA) under the Lifelong Learning Machines (L2M) program Contract No. HR0011-18-C-0051. Any opinions, findings and conclusions or recommendations expressed in this material are those of the authors and do not necessarily reflect the views of the Defense Advanced Research Projects Agency (DARPA).

Special thanks to Roberto Corizzo, Kamil Faber, Nathalie Japkowicz, Michael Baron, James Smith, Sahana Pramod Joshi, Zsolt Kira, Cameron Ethan Taylor, Mustafa Burak Gurbuz, Constantine Dovrolis, Tyler L. Hayes, Christopher Kanan, and Jhair Gallardo for their useful discussions and feedback related to system design.

\bibliography{collas2022_conference}
\bibliographystyle{collas2022_conference}

\appendix
\section{Appendix}

\subsection{Potential Negative and Positive Societal Impacts}
This approach presents a means of training a multi-agent system to play real-time strategy games using lifelong reinforcement learning. Such a tool could be used by malicious foreign actors to train models for strategic military actions, which could negatively impact the national security of benign nations.

However, this tool can also be used for societal benefits including training intelligent robot swarms to perform useful tasks over long lifetimes (e.g., applications in agricultural and manufacturing robotics). Similarly, it could be used in domains which require incorporation of feedback into the learning (i.e. necessitates RL) and where distribution shift is a significant problem; e.g., for planning treatments in healthcare.

\subsection{Limitations}
The model is not flawless. The following are some limitations we have observed w.r.t. our proposed approach:
\begin{itemize}
\item While catastrophic forgetting is reduced using our proposed models, the agent still fails to perfectly learn and remember tasks, and thus, there is still room for improvement in terms of many of the LL metrics. Specifically, we've observed that the model can struggle to remember over very long periods when trained on long syllabi of tasks.
\item Training the agent is non-trivial, requiring carefully defining reasonable tasks and performing hyperparameter tuning. 
\item The sleep model is dependent on having a good quality single-task expert for the wake model, and in some cases, the wake model fails to properly learn a task, disrupting the learning of the sleep model.
\item We have also seen some fragility in the learning of the hidden space for hidden replay and imperfect generated samples in the sequential and two-headed models. We have observed cases where the sampled data drifts enough that it mis-aligns with the true distribution of data and/or the quality of the sampled data degrades significantly. In such cases, the sleep model can become unstable and the proposed replay mechanisms may hurt the model's learning. 
\item While we would like to rely mostly on generative replay and experience replay from the last wake cycle for training the model, we've found that the hidden replay model requires a small buffer of random samples across tasks to maintain the hidden space. This introduces sub-optimality in the memory usage of the model. We have also observed that this ``random replay'' has an out-sized effect on improving lifelong learning  compared to the generative replay.
\end{itemize}

\subsection{Starcraft 2 Task Variants}
\label{sec:sc2_task_descriptions}
The task variants we consider are as follows:

\begin{itemize}
    \item \textbf{Collect Mineral Shards – No Fog of War}: A map with 2 Marines and an endless supply of Mineral Shards. Rewards are earned by moving the Marines to collect the Mineral Shards, with optimal collection requiring both Marine units to be split up and moved independently. Whenever all 20 Mineral Shards have been collected, a new set of 20 Mineral Shards are spawned at random locations (at least 2 units away from all Marines). Fog of war is disabled. 
     \item \textbf{Collect Mineral Shards – Fog of war:} A map with 2 Marines and an endless supply of Mineral Shards. Rewards are earned by moving the Marines to collect the Mineral Shards, with optimal collection requiring both Marine units to be split up and moved independently. Whenever all 20 Mineral Shards have been collected, a new set of 20 Mineral Shards are spawned at random locations (at least 2 units away from all Marines). Fog of war is enabled, meaning the agent must be able to learn without full knowledge of the current state of the environment.
     \item \textbf{DefeatZerglingsAndBanelings – One Group:} A map with 9 Marines on the opposite side from a group of 6 Zerglings and 4 Banelings. Rewards are earned by using the Marines to defeat Zerglings and Banelings. Whenever all Zerglings and Banelings have been defeated, a new group of 6 Zerglings and 4 Banelings is spawned and the player is awarded 4 additional Marines at full health, with all other surviving Marines retaining their existing health (no restore). Whenever new units are spawned, all unit positions are reset to opposite sides of the map
     \item \textbf{DefeatZerglingsAndBanelings – Two Groups:} A map with 9 Marines in the center with two groups consisting of 9 Zerglings one one side and 6 Banelings on the other side. Rewards are earned by using the Marines to defeat Zerglings and Banelings. Whenever a group has been defeated, a new group of 9 Zerglings and 6 Banelings is spawned and the player is awarded 6 additional Marines at full health, with all other surviving Marines retaining their existing health (no restore). Whenever new units are spawned, all unit positions are reset to opposite sides of the map.
     \item \textbf{DefeatRoaches – One Group:} A map with 9 Marines and a group of 4 Roaches on opposite sides. Rewards are earned by using the Marines to defeat Roaches, with optimal combat strategy requiring the Marines to perform focus fire on the Roaches. Whenever all 4 Roaches have been defeated, a new group of 4 Roaches is spawned and the player is awarded 5 additional Marines at full health, with all other surviving Marines retaining their existing health (no restore). Whenever new units are spawned, all unit positions are reset to opposite sides of the map.
     \item \textbf{DefeatRoaches – Two Groups:} A map with 9 Marines in the center and two groups consisting of 6 total Roaches on opposite sides (3 on each side). Rewards are earned by using the Marines to defeat Roaches, with optimal combat strategy requiring the Marines to perform focus fire on the Roaches. Whenever all 6 Roaches have been defeated, a new group of 6 Roaches is spawned and the player is awarded 7 additional Marines at full health, with all other surviving Marines retaining their existing health (no restore). Whenever new units are spawned, all unit positions are reset to starting areas of the map.
\end{itemize}

\subsection{Implementation Details}
\label{exp:archs}

\subsubsection{Starcraft 2}
\label{sec:imp_sc2}

In this section we describe the implementation details for our \SC{} agent and experiments. The wake agent is a VTrace agent \citep{espeholt2018impala}. The feature extractor component of the wake agent consists of an architecture that is a slight modification of the FullyConv model of \citet{vinyals2017starcraft}. We further split this into a feature extractor component and a policy network component to facilitate easy weight copying between the wake and sleep agents. For the hidden replay model, we add a LayerNormalization layer to the output of the feature extractor to constrain the space of the features, which helps to improve feature reconstructions and limit feature drift. The architecture produces features of dimension 8192.

The sleep agent feature extractor and policy network architectures are identical to the wake model. We use a five-layer convolutional neural network architecture with for the encoder and a five-layer deconvolutional architecture for the decoder for the VAEs for the sequential and two-headed models. In both cases, we use ReLU activations between the layers and add batch normalization before the ReLUs. We use two-layered multi-layered perceptrons with ReLU activation functions for the encoder and decoder for the VAE for the hidden replay model. To improve numerical stability, the logvar predicted by the encoder is bounded to take values in $[-5, 5]$ using a scaled-hyperbolic tangent layer.

Three styles of replay are used during sleep for the hidden replay architecture: experience replay on a buffer of 10,000 observation-action pairs from a FIFO queue collected during the wake phase, generative replay on the feature vectors, and random replay. Only experience replay and generative replay is used to train the sleep agent for the sequential and two-headed architectures.

For the sequential architecture: the weight of the imitation loss (cross-entropy on action logits) is 50.0, the weight of the reconstruction loss is 1.0, and the weight of the KL loss of the VAE is 3.0. For the two-headed architecture: the weight of the imitation loss (cross-entropy on action logits) is 30.0, the weight of the reconstruction loss is 1.0, and the weight of the KL loss of the VAE is 2.0. For the hidden replay architecture: the weight of the imitation loss (cross-entropy on action logits) is 50.0, the weight of the reconstruction loss is 200.0, and the weight of the KL loss of the VAE is 5.0. These were determined via hand-tuning.

The agent enters sleep three times per task at even intervals for the sequential and two-headed architectures. The agent enters sleep twice per task at even intervals for the hidden replay architecture. Each training iteration of the wake model is trained on 32 trajectories. Each sleep consists of 4,500 iterations of training using batch sizes of 64 (64 wake samples, 64 generated samples, and 64 random observations from the random replay buffer). Generative replay is not used until the second sleep. 96 observations are selected randomly from the wake buffer every sleep to be added to the random replay buffer for the hidden replay architecture. When the agent wakes up, it resets the PPO model and offers advice with a 80\% probability linearly decaying to 0\% after for one-half the the duration between sleeps. No weight copying is used for the sequential variant, but weight copying (running a mini-evaluation block after sleep and copying the best Eigentask's parameters into the wake agent) and advice are both used for the two-headed and hidden replay variants.

The Adam optimizer is used to train both the wake and sleep agents with a learning rate of 1.0e-3. 

\textbf{Evaluation:} We selected 3 minigames involving battles  (\emph{DefeatRoaches} and \emph{DefeatZerglingsAndBanelings} tasks) or resource collection (\emph{CollectMineralShards}). We created two variants of each task, differentiated by the starting locations of units, the presence of ``fog-of-war'', and/or the number of friendly and enemy units. Task variants are described in Section~\ref{sec:sc2_task_descriptions}. We used PySC2 \citep{vinyals2017starcraft} to interface with SC-2. We used a subset of the available observation maps namely the unit type, selection status, and unit density two-dimensional observations. The action space is factored into functions and arguments, such as $\text{move}(x, y)$ or $\text{stop}()$. Following \citet{vinyals2017starcraft}, our policy networks output probabilities for the different action factors independently. 
The agent receives positive rewards for collecting resources and defeating enemy units, and negative rewards for losing friendly units. Single task experts (STEs) were trained to convergence ($\sim$ 30 million environment steps using the VTrace algorithm \citep{espeholt2018impala}). We consider the pairwise, alternating, and condensed scenarios where each LB consists of 2 million environment steps (about 6\% of the STE samples), and each EB consists of 30 episodes per task.

Each lifetime of a condensed scenario takes approximately 3-4 days running on systems with RTX2080 GPUs. 


\begin{figure}[H]
\centering
\includegraphics[width=0.95\textwidth]{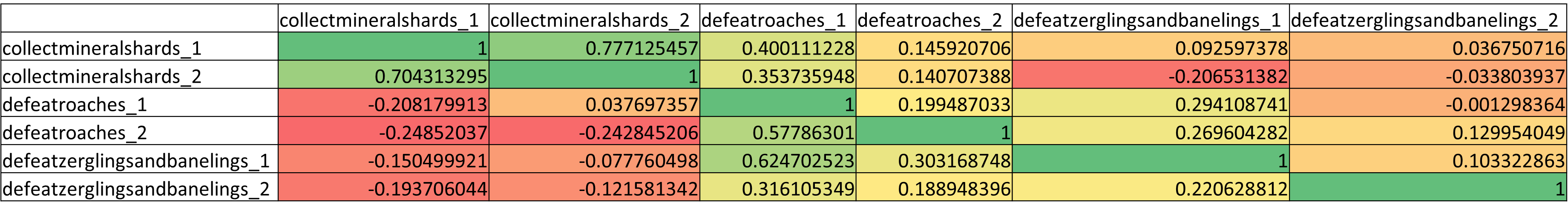}
\caption{Similarity between tasks measured in terms of forward transfer using single task experts}
\label{fig:similarity_matrix}
\end{figure}

\subsubsection{Minigrid}
\label{sec:imp_minigrid}
In this section, we describe the implementation details for our Minigrid agent and experiments. The wake agent is based on the Stable Baselines 3 PPO agent \citep{schulman2017proximal,stable-baselines3}. There are separate networks for computing the value and policy. Both networks are simple multi-layered perceptrons that first map to a 256-dimensional feature space and then apply a densely connected layer to select one of six actions or estimate the value of a state. ReLUs are used as activation functions between all of the layers. Observations are scaled so the values of every cell are in $[0, 1]$ before being input to the network. For exact parameters for the wake agent, please refer to the appendix.

The sleep agent uses one Eigentask. The feature extractor and expected inputs are the same as those of the wake agent's policy network. A variational autoencoder is used as generative model. The VAE is a four-layer bottleneck-style MLP architecture with a latent dimension size: 128. To improve numerical stability, the logvar predicted by the encoder is bounded to take values in $[-5, 5]$.

Three styles of replay are used during sleep: experience replay on a buffer of 20,000 observation-action pairs from a FIFO queue collected during the wake phase, generative replay on the feature vectors, and random replay (256 random observations added per sleep). The weight of the imitation loss (cross-entropy on discrete actions) is 3.0, the weight of the reconstruction loss is 1.0, and the weight of the KL loss of the VAE is 0.03. The agent enters sleep at the end of each task. Each sleep consists of 20,000 iterations of training using batch sizes of 32 (32 wake samples, 32 generated samples, and 32 random examples drawn from the random replay buffer). Generative replay and random replay are not used until the second sleep. 256 observations are selected randomly from the wake buffer every sleep to be added to the random replay buffer with a maximum buffer size of 4096. When the agent wakes up, it resets the PPO model and offers advice with a 90\% probability linearly decaying to 0\% after 100,000 environment steps for the condensed scenario and 33,333 environment steps for the dispersed scenario. No weight copying is used.

The Adam optimizer is used to train both the wake and sleep agents.

\textbf{Evaluation:} We report results using the Minigrid domain \citep{gym_minigrid}, in which an agent must navigate to a goal while interacting with various entities in the environment. We consider five tasks --- SimpleCrossing, DistShift, a custom Fetch, a custom Unlock, and DoorKey --- each with two variants (differing in the size of the grid and/or number of objects) for a total of ten POMDPs. Descriptions of these tasks can be found in \citet{gym_minigrid}. An observation consists of a $7\times7$ top-down view of the grid directly in front of the agent, with three channels containing information about the object type, object color, and object state in each tile. Rewards are sparse and consist of $-1$ for running into obstacles and lava, $0$ for not reaching the goal, and a positive reward in $[0, 1]$ for reaching the goal that is proportional to the number of steps taken. Unlike SC-2, there are only six actions: turn left, turn right, move forward, pick up an object, drop an object, and interact with an object.
STEs are trained on each task for 1M environment steps using PPO \citep{schulman2017proximal}. We consider the pairwise, alternating, and condensed scenarios where the LB lengths vary between 100k and 700k (Section~\ref{sec:minigrid_task_lengths}) based on examining the smallest number of steps for the STE to converge to $80\%$ of its peak performance averaged over ten runs. Each EB consists of 100 episodes per task.

Each lifetime of a condensed scenario takes approximately 4-6 hours to run on systems with RTX2080 GPUs. 

\subsection{Minigrid Task Lengths}
\label{sec:minigrid_task_lengths}
We use the following number of environment steps per task in our minigrid experiments:
\begin{itemize}
    \item SimpleCrossingS9N1: 400,000
    \item SimpleCrossingS9N2: 500,000
    \item DistShiftR2: 200,000
    \item DistShiftR3: 200,000
    \item CustomFetchS5T1N2: 500,000
    \item CustomFetchS8T1N2: 700,000
    \item CustomUnlockS5: 200,000
    \item CustomUnlockS7: 300,000
    \item DoorKeyS5: 200,000
    \item DoorKeyS6: 300,000
\end{itemize}

\subsection{Parameters of the PPO agent}
These are the parameters used for the stable baselines 3 PPO agent for the Minigrid tasks:
\begin{itemize}
    \item n\_steps: 512
    \item batch\_size: 32
    \item gae\_lambda: 0.95
    \item gamma: 0.99
    \item n\_epochs: 10
    \item ent\_coef: 5.0e-5
    \item learning\_rate: 2.5e-4
    \item clip\_range: 0.3
    \item vf\_coef: 0.75
    \item max\_grad\_norm: 5.0
    \item policy\_kwargs: dict(activation\_fn=nn.ReLU, net\_arch=[dict(pi=[256, 256], vf=[256, 256])])
\end{itemize}

\subsection{Evaluation Scenarios Visualization}

\begin{figure}[H]
\centering
    \includegraphics[width=0.75\textwidth, keepaspectratio]{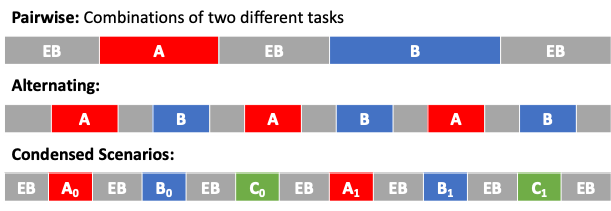}
  \caption{Lifelong Learning Scenarios. Grey evaluation blocks (EBs) evaluate the agent on all tasks. Learning blocks (LBs) are color coded by task, and subscripts denote task variants.}
  \label{fig:scenarios}
\end{figure}

\subsection{Example Learning Curves for Lifelong Curricula on \SC{}}

\begin{figure}[H]
\centering
\includegraphics[width=0.95\textwidth]{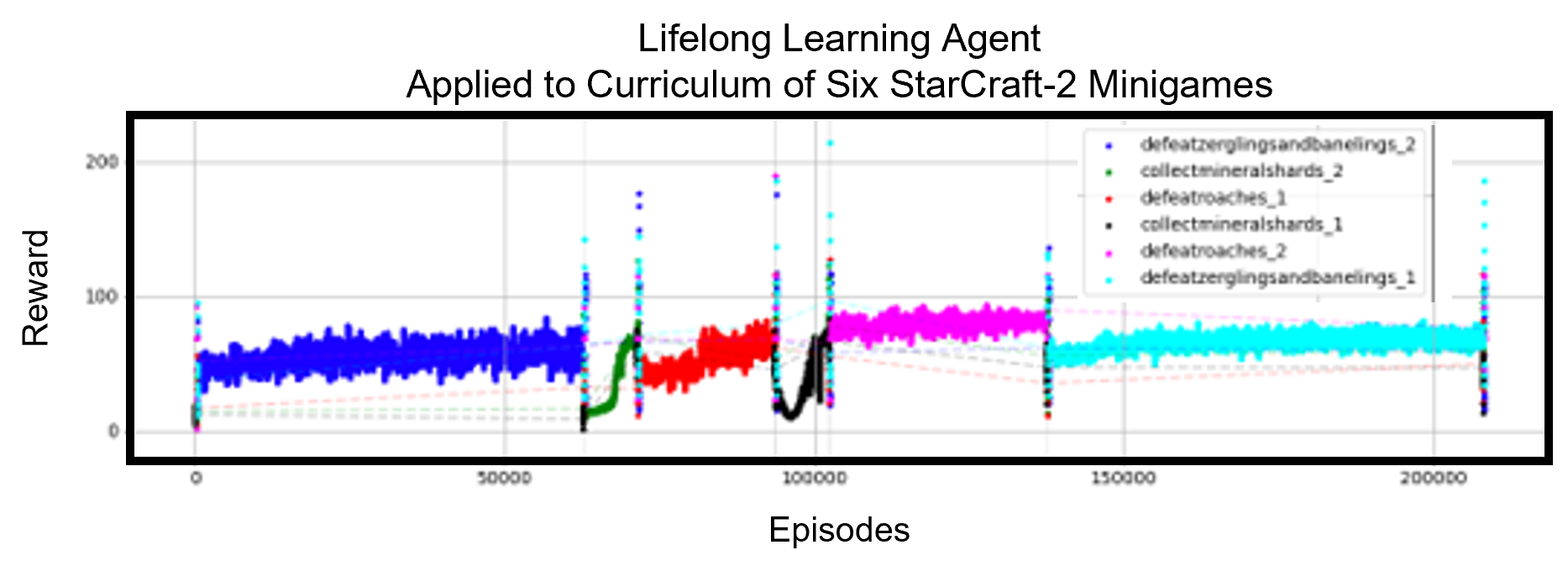}
\caption{Learning curve for our lifelong learning agent applied to a six minigame scenario in the \SC{} domain. Performance comparable to the converged single-task expert is \textasciitilde 100.}
\label{fig:sc2_learning_curves}
\end{figure}

\subsection{Comparison of Learning Curves for Lifelong Curricula on Minigrid}

\begin{figure}[H]
\centering
\includegraphics[width=0.45\textwidth]{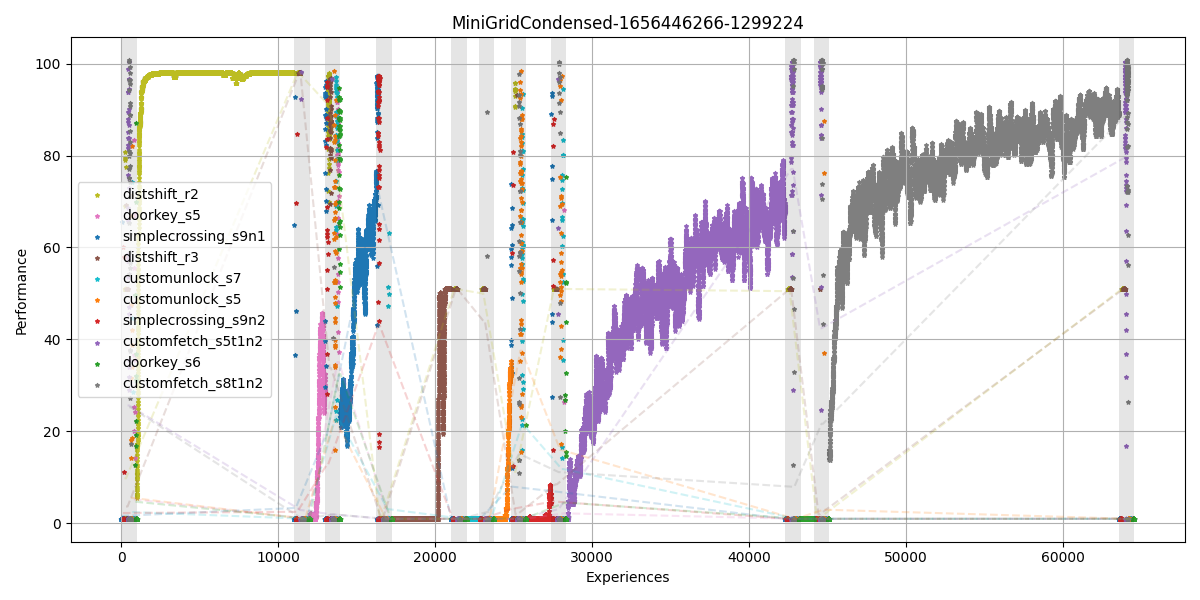}
\includegraphics[width=0.45\textwidth]{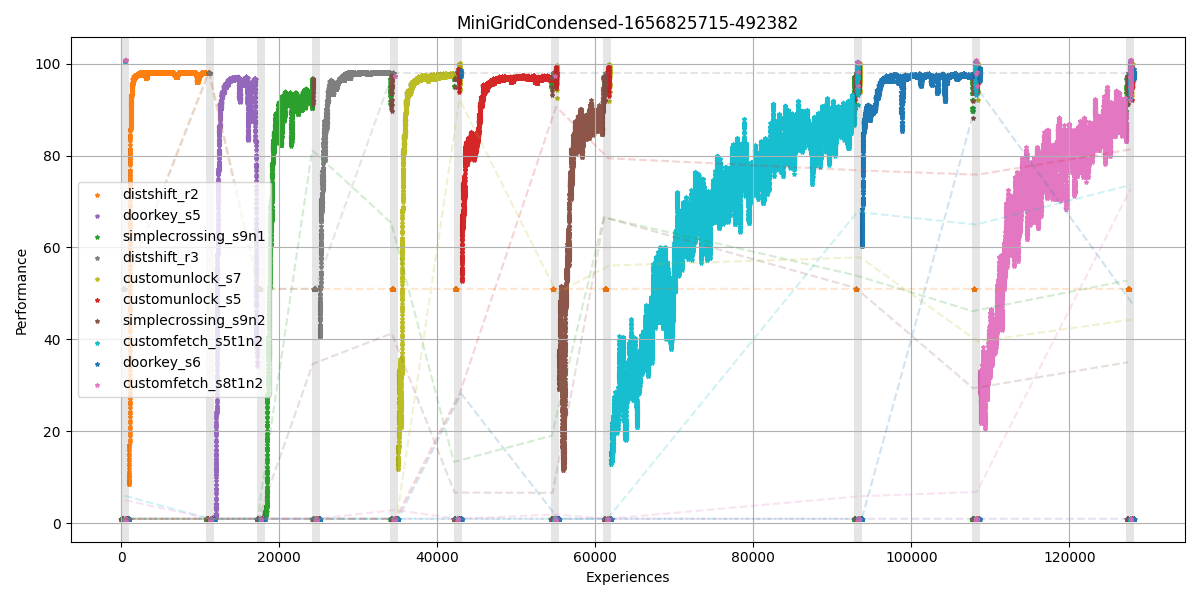}
\caption{Learning curves for a condensed scenario in the Minigrid domain. Left: learning curve of non-LL agent; Right: learning curve of LL agent on the same scenario. Maximum reward is 100. The LL agent is capable of learning more tasks without interference and exhibits jump starts in performance for similar tasks due to the advice mechanism.}
\label{fig:minigrid_learning_curves}
\end{figure}

\subsection{A Sleep Policy Trained with Policy Distillation can Learn to Imitate a Converged Single Task Expert}
\label{sec:ste_additional_results}

\begin{figure}[H]
\centering
\includegraphics[width=0.95\textwidth]{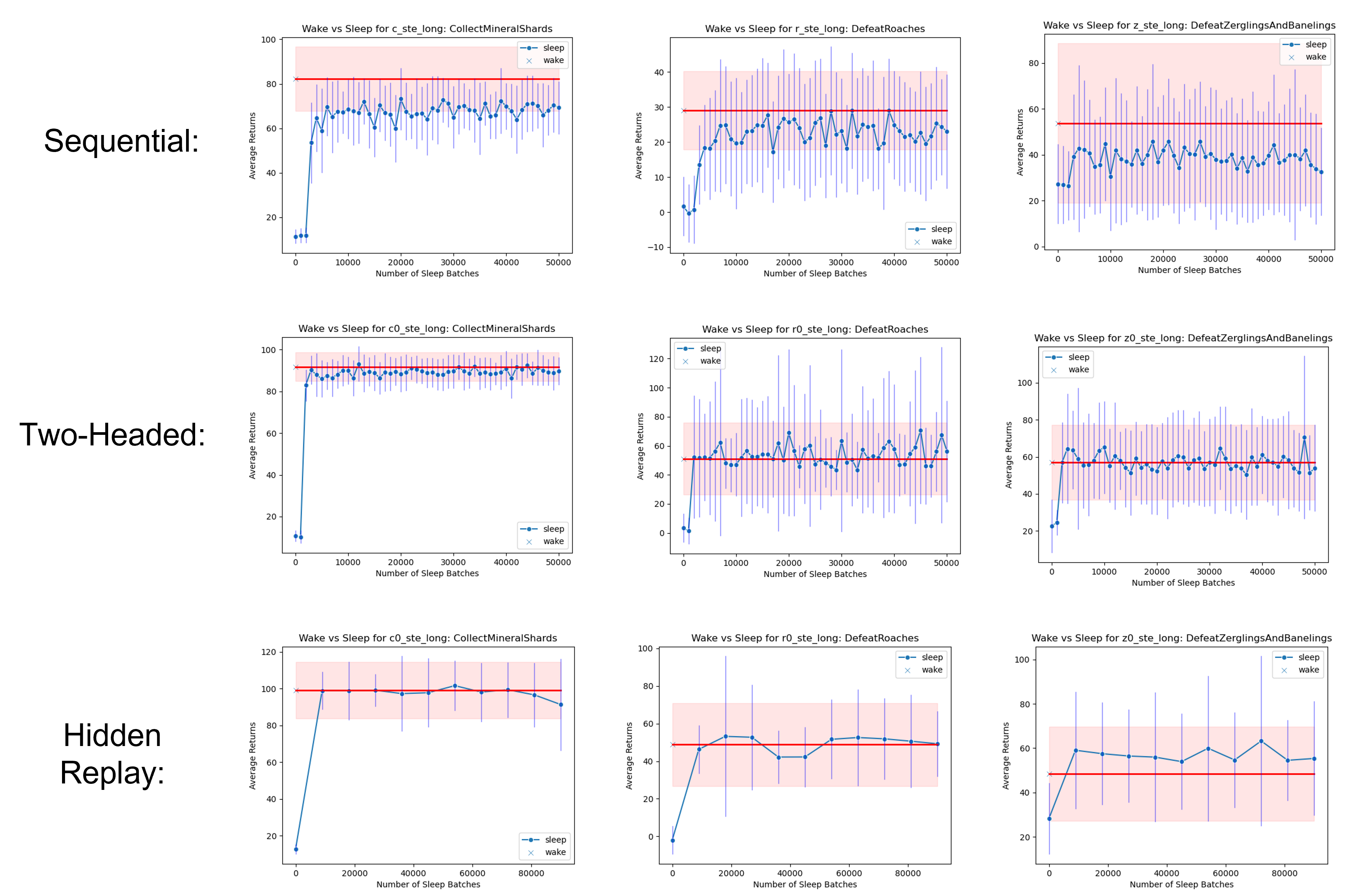}
\caption{Examining whether sleep policies can be trained to mimic the performance of single task experts for the SC-2 POMDPs with the sequential, two-headed, and hidden replay model without a random replay buffer. The blue curves denote the performance (in terms of rewards) of the sleep policy interpolated along evaluation points. The red line denotes terminal performance of the STE after convergence. Error bars denote standard deviation of rewards.}
\label{fig:sleep_imitating_ste}
\end{figure}

\begin{figure}[H]
\centering
\includegraphics[width=0.95\textwidth]{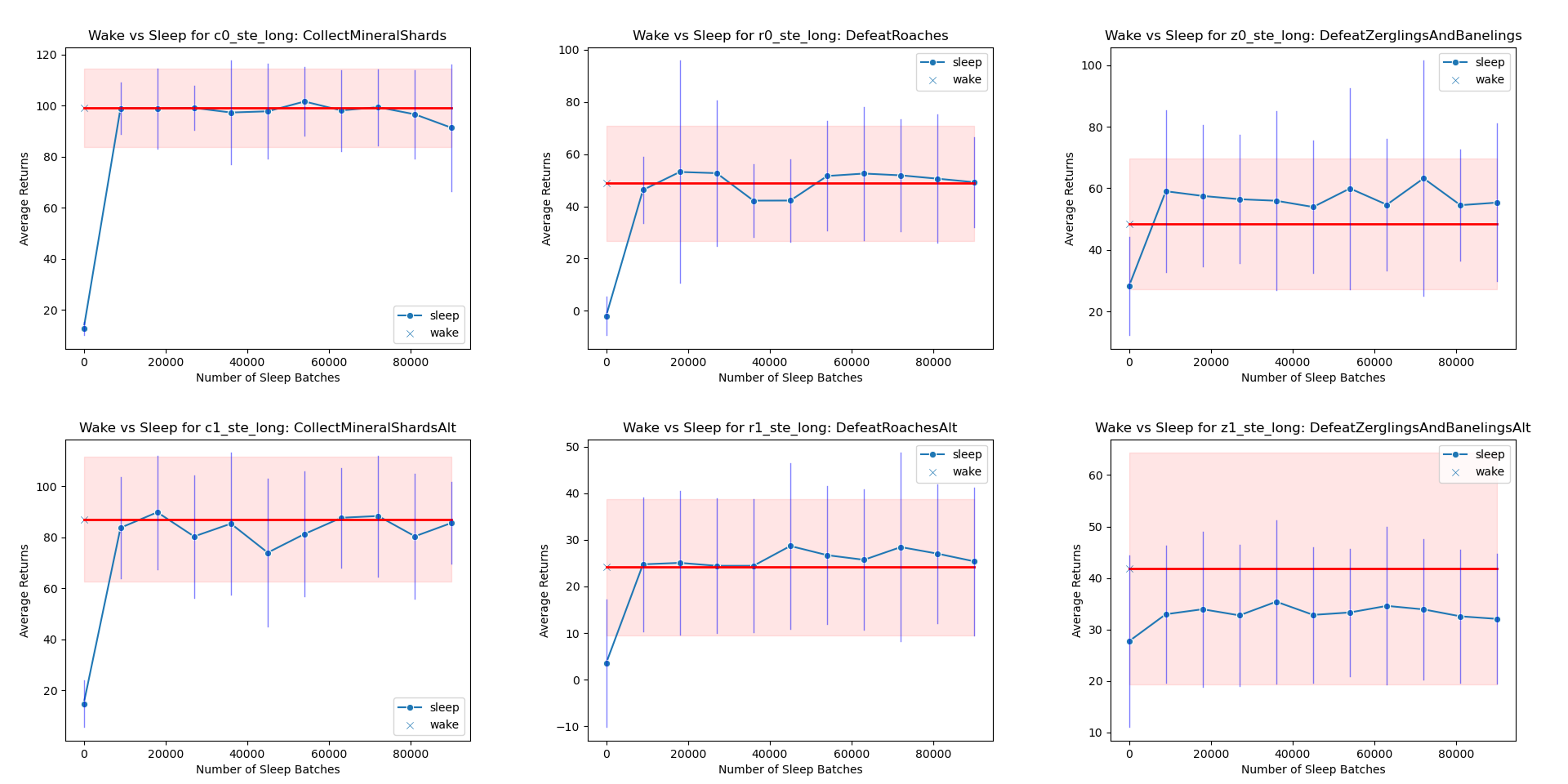}
\caption{Examining whether the sleep policy can be trained to mimic the performance of single task experts for the SC-2 POMDPs with the hidden replay model without a random replay buffer. The blue curves denote the performance (in terms of rewards) of the sleep policy interpolated along evaluation points. The red line denotes terminal performance of the STE after convergence. Error bars denote standard deviation of rewards.}
\label{fig:sleep_imitating_ste_app}
\end{figure}

\textbf{Single Task Experiments}: 


\textbf{SC-2:} We demonstrate that a sleep policy can be trained with policy distillation to mimic the performance of an STE's policy. We consider STEs identical to the wake agents trained for 30M environment steps for each of the six SC-2 variants. We act in accordance with the following procedure: i) Train an STE on each of the six SC-2 POMDPs for 30M steps. ii)  Measure the final reward for each STE. iii) Load the trained STE and build a buffer of 10K trajectories. iv) Train the sleep model until convergence on randomly sampled batches of observation-action pairs (batch size: 32). v) Periodically evaluate the sleep model on SC-2 simulations, each consisting of 30 episodes of the task-of-interest.
We show results in Figs.~\ref{fig:sleep_imitating_ste} and \ref{fig:sleep_imitating_ste_app}. The two-headed and hidden replay architectures architectures typically match or exceed the STE performance in terms of rewards. The sequential model learns non-trivial policies, but generally performs worse than the STE. This supports our hypothesis that decoupling the generator from the policy network of the VAE results in better-performing policies.

\textbf{Minigrid:} We also quantitatively validate that the sleep policy using the hidden replay model is able to distill the an STE's converged policy after 1M environment steps for the Minigrid domain using an experience buffer of the STE's last 20K observation-action pairs (repeated 10 times per task). We find that the average difference in rewards between the STE and sleep policy's performance over all task variants is 0.14 with a standard deviation of 0.11, suggesting that the sleep policy imitates the STE to an acceptable level.

\subsection{A Sleep Policy Trained with Policy Distillation can Learn to Imitate the Wake Policy on a Single Task}
\label{sec:wake_additional_results}

We conduct experiments to show that the converged performance of a wake policy can be achieved by our agent via imitating the sequence of intermediate wake policies over multiple wake-sleep phases. This is an important experiment needed to show stable convergence of the wake-sleep mechanism even when distilling sequences of sub-optimal policies. These experiments involve training an agent on a single task, periodically entering sleep at set intervals. We observed that for most tasks, the curves formed by evaluating the sleep agent and wake agents at periodic evaluation blocks closely track one another, demonstrating stable convergence of the sleep agent's policy.

\begin{figure}[H]
\centering
\includegraphics[width=0.95\textwidth]{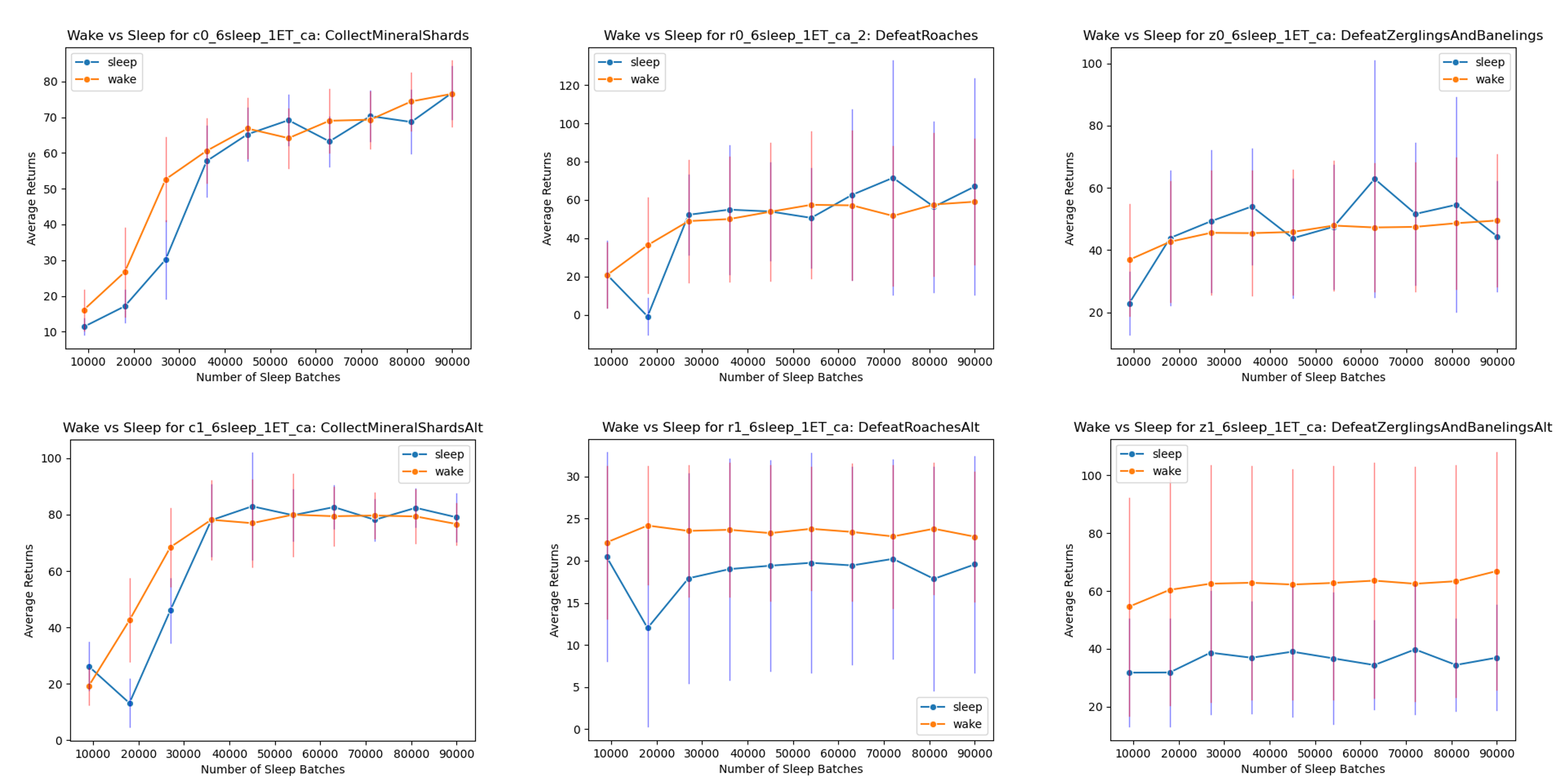}
\caption{Examining whether the sleep policy can be trained to mimic the performance of wake non-lifelong learner for the SC-2 POMDPs as the non-lifelong learners improve. This experiment examined the performance of the hidden replay model without using observations from the random replay buffer. The blue curves denote the performance (in terms of rewards) of the sleep policy interpolated along evaluation points. The red line denotes performance of the wake policy interpolated along evaluation points. Error bars denote standard deviation of rewards.}
\label{fig:sleep_imitating_wake_app}
\end{figure}

\textbf{SC-2:} We consider the case where the system performs multiple wake-sleep phases over the course of a single task, requiring the sleep policy to consolidate the increasingly improving policies of the wake agent. Using the SC-2 simulator, each wake policy is trained for 6M environment steps, entering sleep at even intervals ten times over the entire run. Every time the system exits sleep, the wake and sleep policies are evaluated on thirty episodes of the task. We show the wake versus sleep performance (in terms of rewards) of a hidden replay-based sleep policy in Fig.\ \ref{fig:sleep_imitating_wake_app}. The sleep policy's performance typically closely mirrors the wake policy's performance, but there are cases where the agent learns a non-trivial policy, but under-performs the wake policy. While not pictured here, similar behavior was observed for the sequential and two-headed architectures as well.


\textbf{Minigrid:} We also conducted a post-hoc analysis of our condensed scenario runs in the Minigrid domain to quantify the sleep policy's capability to imitate the wake policy after learning the first task. 

\textit{\textbf{Sequential Model:}} The mean absolute difference between the wake policy's average reward for the last 100 episodes and the sleep policy's average reward for the evaluation block was 0.589 with a standard deviation of 0.320, suggesting the sleep policy when using the sequential architecture struggles to accurately imitate the wake policy in many, but not all runs. In cases where the wake model converges to a policy with average reward $>0.9$ for the last 100 episodes, the mean difference metric decreases to $0.591 \pm 0.384$.

\textit{\textbf{Two-Headed Model:}} The mean absolute difference between the wake policy's average reward for the last 100 episodes and the sleep policy's average reward for the evaluation block was 0.422 with a standard deviation of 0.263, suggesting the sleep policy when using the two-headed model can struggle to accurately imitate the wake policy in some runs. In cases where the wake model converges to a policy with average reward $>0.9$ for the last 100 episodes, the mean difference metric decreases to $0.341 \pm 0.297$.

\textit{\textbf{Hidden Replay Model:}} The mean absolute difference between the wake policy's average reward for the last 100 episodes and the sleep policy's average reward for the evaluation block was 0.209 with a standard deviation of 0.227, suggesting the sleep policy accurately imitates the wake policy in many, but not all runs. In cases where the wake model converges to a policy with average reward $>0.9$ for the last 100 episodes, the mean difference metric decreases to $0.110 \pm 0.145$.

These results suggest that the hidden replay model has an advantage when it comes to imitating a wake model when learning a task from scratch. The sequential model struggles with imitating the wake model when learning a task from scratch; however, our main results in the paper suggest that as the sequential model encounters additional tasks, this issue of successfully imitating the wake policy is somewhat alleviated as the relative reward post learning block metric ultimately is similar to the two-headed model when deployed over an entire syllabus of tasks.



\subsection{Effect of Varying the Number of Wake-Sleep Cycles}
we consider the case where the system encounters multiple tasks in sequence, and we evaluate the effect of the number of wake-sleep phases per task on the lifelong learning metrics. We trained the two-headed replay architecture for 4 lifetimes, each consisting of alternating between an \SC{} combat and collection task with 2M environment steps per task. We compared the effect of one, two, and three sleeps per task and report results in Table \ref{tab:multi-sleep}. Results suggest that multiple sleep phases result in improved or maintained performance in most metrics, but may negatively affect FT.


\begin{table}[H]
    \centering
    \resizebox{1.0\textwidth}{!}{
    \begin{tabular}{|c|c|c|c|c||c|c|c|c|c|}
    \hline
    Agent & PM & FT & BT & RP & RR\textsubscript{$\Omega$} & RR\textsubscript{$\alpha$} & RR\textsubscript{$\sigma$} & RR\textsubscript{$\upsilon$} \\
    \hline
One Sleep Cycle Per Task & $-8.76$ ($ \pm 4.39$) & $1.68$ ($ \pm 2.16$) & $0.76$ ($ \pm 0.06$) & $0.78$ ($ \pm 0.22$) & $0.58$ ($ \pm 0.29$) & $0.7$ ($ \pm 0.22$) & $0.58$ ($ \pm 0.22$) & $0.3$ ($ \pm 0.38$)\\
Two Sleep Cycles Per Task & $-10.47$ ($ \pm 6.41$) & $1.16$ ($ \pm 0.67$) & $0.93$ ($ \pm 0.11$) & $0.87$ ($ \pm 0.11$) & $0.79$ ($ \pm 0.47$) & $0.95$ ($ \pm 0.25$) & $0.81$ ($ \pm 0.31$) & $0.27$ ($ \pm 0.31$)\\
Three Sleep Cycles Per Task & $-8.63$ ($ \pm 7.13$) & $1.06$ ($ \pm 0.86$) & $0.78$ ($ \pm 0.25$) & $0.84$ ($ \pm 0.13$) & $0.67$ ($ \pm 0.19$) & $0.7$ ($ \pm 0.13$) & $0.61$ ($ \pm 0.1$) & $0.19$ ($ \pm 0.19$)\\
\hline
    \end{tabular}
    }
    \caption{Comparing the effect of sleeping multiple times per task on a subset of four alternating scenarios involving transitioning between combat and collection \SC{} tasks. We see in many cases that sleeping multiple times per task can improve metrics.}
    \label{tab:multi-sleep}
\end{table}

\subsection{Understanding the Effect of Different Replay Types}
\label{app:replay_ablation}

In Tables \ref{tab:replay_mechanism_ablation_minigrid} and \ref{tab:replay_mechanism_ablation_sc2}, we investigate the effects of turning on and off different replay mechanisms in the hidden replay model. In general, we validate that combining experience replay of the wake buffer, generative replay, and random replay, results in an improved agent; although experience replay in combination with random replay performs surprisingly well.

\begin{table}[H]
    \centering
    \resizebox{1.0\textwidth}{!}{
    \begin{tabular}{|c|c|c|c|c|c||c|c|c|c|}
    \hline
    Scenario & Agent & PM & FT & BT & RP & RR\textsubscript{$\Omega$} & RR\textsubscript{$\alpha$} & RR\textsubscript{$\sigma$} & RR\textsubscript{$\upsilon$} \\
    \hline
    & Random Policy & $0.37$ ($ \pm 0.81$)  & $1.06$ ($ \pm 0.02$)  & $1.13$ ($ \pm 0.18$)  & $0.22$ ($ \pm 0.04$)  & $0.09$ ($ \pm 0.02$)  & $0.09$ ($ \pm 0.02$)  & $0.09$ ($ \pm 0.02$)  & $0.09$ ($ \pm 0.00$) \\
& Baseline PPO & $-48.2$ ($ \pm 12.04$) & $4.07$ ($ \pm 0.87$) & $0.46$ ($ \pm 0.12$) & $2.17$ ($ \pm 0.61$) & $0.5$ ($ \pm 0.1$) & $0.78$ ($ \pm 0.08$) & $0.64$ ($ \pm 0.08$) & $0.16$ ($ \pm 0.02$)\\
Pairwise & Hidden Replay (ER) & $-45.25$ ($ \pm 13.54$) & $6.34$ ($ \pm 1.4$) & $0.51$ ($ \pm 0.26$) & $2.75$ ($ \pm 0.73$) & $0.47$ ($ \pm 0.06$) & $0.73$ ($ \pm 0.08$) & $0.6$ ($ \pm 0.06$) & $0.11$ ($ \pm 0.02$) \\
& Hidden Replay (ER + RaR)  & $-21.11$ ($ \pm 11.06$) & $5.97$ ($ \pm 1.84$) & $0.75$ ($ \pm 0.19$) & $2.22$ ($ \pm 0.91$) & $0.56$ ($ \pm 0.08$) & $0.68$ ($ \pm 0.08$) & $0.62$ ($ \pm 0.08$) & $0.11$ ($ \pm 0.02$)\\
& Hidden Replay (ER + GR) & $-62.32$ ($ \pm 25.2$) & $7.12$ ($ \pm 2.51$) & $0.31$ ($ \pm 0.29$) & $2.51$ ($ \pm 1.61$) & $0.4$ ($ \pm 0.11$) & $0.74$ ($ \pm 0.13$) & $0.57$ ($ \pm 0.11$) & $0.11$ ($ \pm 0.02$)\\
& Hidden Replay (ER + RaR + GR) & $-15.77$ ($ \pm 10.09$) & $5.87$ ($ \pm 1.24$) & $1.21$ ($ \pm 0.55$) & $2.47$ ($ \pm 0.69$) & $0.6$ ($ \pm 0.06$) & $0.69$ ($ \pm 0.08$) & $0.64$ ($ \pm 0.06$) & $0.11$ ($ \pm 0.02$) \\
\hline
& Random Policy & $-0.02$ ($ \pm 0.26$)  & $1.08$ ($ \pm 0.02$)  & $1.08$ ($ \pm 0.02$)  & $0.19$ ($ \pm 0.00$)  & $0.09$ ($ \pm 0.00$)  & $0.09$ ($ \pm 0.00$)  & $0.09$ ($ \pm 0.00$)  & $0.09$ ($ \pm 0.00$) \\
& Baseline PPO & $-38.92$ ($ \pm 3.98$) & $4.72$ ($ \pm 0.75$) & $4.06$ ($ \pm 0.73$) & $1.44$ ($ \pm 0.2$) & $0.22$ ($ \pm 0.04$) & $0.64$ ($ \pm 0.04$) & $0.33$ ($ \pm 0.04$) & $0.15$ ($ \pm 0.02$)\\
Condensed & Hidden Replay (ER) & $-57.71$ ($ \pm 3.61$) & $6.45$ ($ \pm 0.81$) & $5.32$ ($ \pm 0.81$) & $2.04$ ($ \pm 0.1$)  & $0.17$ ($ \pm 0.02$) & $0.77$ ($ \pm 0.04$)  & $0.33$ ($ \pm 0.02$) & $0.12$ ($ \pm 0.02$)\\
& Hidden Replay (ER + RaR) & $-20.96$ ($ \pm 2.81$) & $5.44$ ($ \pm 0.93$) & $1.85$ ($ \pm 0.45$) & $2.11$ ($ \pm 0.1$) & $0.54$ ($ \pm 0.04$) & $0.77$ ($ \pm 0.02$) & $0.61$ ($ \pm 0.02$) & $0.16$ ($ \pm 0.02$)\\
& Hidden Replay (ER + GR) & $-52.91$ ($ \pm 7.26$) & $6.75$ ($ \pm 1.03$) & $4.93$ ($ \pm 1.8$) & $1.97$ ($ \pm 0.24$) & $0.19$ ($ \pm 0.02$) & $0.75$ ($ \pm 0.07$) & $0.33$ ($ \pm 0.02$) & $0.12$ ($ \pm 0.02$)\\
& Hidden Replay (ER + RaR + GR) & $-21.94$ ($ \pm 2.09$) & $5.93$ ($ \pm 0.79$) & $1.32$ ($ \pm 0.2$)  & $2.1$ ($ \pm 0.1$)  & $0.56$ ($ \pm 0.04$)  & $0.8$ ($ \pm 0.02$)  & $0.63$ ($ \pm 0.02$)  & $0.16$ ($ \pm 0.02$)\\
    \hline
    \end{tabular}
    }
    \caption{Effect of different replay mechanisms during sleep on lifelong learning metrics on the Minigrid domain. Results are collected using the hidden replay model on thirty-six syllabi. We report means and 95\% confidence intervals. Hidden replay using only experience replay (ER) and generative replay (GR) is insufficient. Random replay (RaR) is needed to stablize the learning of the feature space, and combining all three types of replay yield the best results.}
    \label{tab:replay_mechanism_ablation_minigrid}
\end{table}

\begin{table}[H]
    \centering
    \resizebox{1.0\textwidth}{!}{
    \begin{tabular}{|c|c|c|c|c|c||c|c|c|c|}
    \hline
    Scenario & Agent & PM & FT & BT & RP & RR\textsubscript{$\Omega$} & RR\textsubscript{$\alpha$} & RR\textsubscript{$\sigma$} & RR\textsubscript{$\upsilon$} \\
    \hline
    & Random Policy & $0.0$ ($ \pm 0.00$) & $1.0$ ($ \pm 0.00$) & $1.0$ ($ \pm 0.00$) & $0.42$ ($ \pm 0.04$) & $0.32$ ($ \pm 0.09$) & $0.32$ ($ \pm 0.09$) & $0.32$ ($ \pm 0.09$) & $0.32$ ($ \pm 0.22$)\\
& Baseline VTrace & $-8.99$ ($ \pm 3.06$) & $1.11$ ($ \pm 0.23$) & $0.79$ ($ \pm 0.12$) & $0.92$ ($ \pm 0.04$) & $0.9$ ($ \pm 0.14$) & $0.91$ ($ \pm 0.12$) & $0.82$ ($ \pm 0.12$) & $0.39$ ($ \pm 0.17$)\\
    & Hidden Replay(ER) & $-7.9$ ($ \pm 5.11$) & $1.14$ ($ \pm 0.4$) & $0.75$ ($ \pm 0.29$) & $0.92$ ($ \pm 0.09$) & $0.87$ ($ \pm 0.11$) & $0.9$ ($ \pm 0.11$) & $0.8$ ($ \pm 0.13$) & $0.37$ ($ \pm 0.2$)\\
Pairwise & Hidden Replay(ER $+$ RaR) & $-3.7$ ($ \pm 4.18$) & $1.75$ ($ \pm 0.66$) & $0.92$ ($ \pm 0.11$) & $1.07$ ($ \pm 0.11$) & $0.78$ ($ \pm 0.18$) & $0.81$ ($ \pm 0.18$) & $0.79$ ($ \pm 0.18$) & $0.48$ ($ \pm 0.26$)\\
& Hidden Replay(ER $+$ GR) & $-9.35$ ($ \pm 7.46$) & $1.39$ ($ \pm 0.64$) & $0.71$ ($ \pm 0.18$) & $1.05$ ($ \pm 0.09$) & $0.71$ ($ \pm 0.15$) & $0.82$ ($ \pm 0.11$) & $0.77$ ($ \pm 0.13$) & $0.44$ ($ \pm 0.31$)\\
& Hidden Replay (ER $+$ RaR $+$ GR) & $-0.57$ ($ \pm 2.93$) & $1.85$ ($ \pm 0.88$) & $0.95$ ($ \pm 0.15$) & $1.04$ ($ \pm 0.11$) & $0.77$ ($ \pm 0.18$) & $0.82$ ($ \pm 0.13$) & $0.8$ ($ \pm 0.15$) & $0.55$ ($ \pm 0.26$)\\
\hline
& Random Policy & $0.0$ ($ \pm 0.00$) & $1.0$ ($ \pm 0.00$) & $1.0$ ($ \pm 0.00$) & $0.42$ ($ \pm 0.04$) & $0.32$ ($ \pm 0.09$) & $0.32$ ($ \pm 0.09$) & $0.32$ ($ \pm 0.09$) & $0.32$ ($ \pm 0.22$)\\
& Baseline VTrace & $-8.99$ ($ \pm 3.06$) & $1.11$ ($ \pm 0.23$) & $0.79$ ($ \pm 0.12$) & $0.92$ ($ \pm 0.04$) & $0.9$ ($ \pm 0.14$) & $0.91$ ($ \pm 0.12$) & $0.82$ ($ \pm 0.12$) & $0.39$ ($ \pm 0.17$)\\
Alternating & Hidden Replay(ER) & $-7.9$ ($ \pm 5.11$) & $1.14$ ($ \pm 0.4$) & $0.75$ ($ \pm 0.29$) & $0.92$ ($ \pm 0.09$) & $0.87$ ($ \pm 0.11$) & $0.9$ ($ \pm 0.11$) & $0.8$ ($ \pm 0.13$) & $0.37$ ($ \pm 0.2$)\\
& Hidden Replay(ER $+$ RaR) & $-4.25$ ($ \pm 2.77$) & $1.75$ ($ \pm 0.66$) & $0.88$ ($ \pm 0.11$) & $0.91$ ($ \pm 0.09$) & $0.81$ ($ \pm 0.15$) & $0.83$ ($ \pm 0.13$) & $0.79$ ($ \pm 0.15$) & $0.48$ ($ \pm 0.26$)\\
& Hidden Replay(ER $+$ GR) & $-10.99$ ($ \pm 4.89$) & $1.39$ ($ \pm 0.64$) & $0.65$ ($ \pm 0.15$) & $0.87$ ($ \pm 0.09$) & $0.83$ ($ \pm 0.18$) & $0.89$ ($ \pm 0.11$) & $0.76$ ($ \pm 0.13$) & $0.44$ ($ \pm 0.31$)\\
& Hidden Replay (ER $+$ RaR $+$ GR) & $-6.13$ ($ \pm 4.64$) & $1.85$ ($ \pm 0.88$) & $0.87$ ($ \pm 0.13$) & $0.91$ ($ \pm 0.09$) & $0.86$ ($ \pm 0.13$) & $0.89$ ($ \pm 0.13$) & $0.82$ ($ \pm 0.15$) & $0.55$ ($ \pm 0.26$)\\
\hline
& Random Policy & $-0.03$ ($ \pm 0.28$) & $1.01$ ($ \pm 0.02$) & $1.01$ ($ \pm 0.02$) & $0.54$ ($ \pm 0.02$) & $0.31$ ($ \pm 0.02$) & $0.3$ ($ \pm 0.02$) & $0.31$ ($ \pm 0.00$) & $0.3$ ($ \pm 0.04$)\\
& Baseline VTrace & $-3.41$ ($ \pm 1.7$) & $1.19$ ($ \pm 0.08$) & $1.17$ ($ \pm 0.29$) & $1.14$ ($ \pm 0.06$) & $0.64$ ($ \pm 0.06$) & $0.74$ ($ \pm 0.06$) & $0.66$ ($ \pm 0.06$) & $0.49$ ($ \pm 0.06$)\\
Condensed & Hidden Replay(ER) & $-8.65$ ($ \pm 3.15$) & $1.42$ ($ \pm 0.15$) & $1.12$ ($ \pm 0.11$) & $1.12$ ($ \pm 0.09$) & $0.73$ ($ \pm 0.09$) & $0.89$ ($ \pm 0.07$) & $0.77$ ($ \pm 0.04$) & $0.55$ ($ \pm 0.07$)\\
& Hidden Replay(ER $+$ RaR) & $-3.53$ ($ \pm 1.67$) & $1.35$ ($ \pm 0.09$) & $1.07$ ($ \pm 0.13$) & $1.19$ ($ \pm 0.07$) & $0.8$ ($ \pm 0.04$) & $0.9$ ($ \pm 0.04$) & $0.81$ ($ \pm 0.04$) & $0.61$ ($ \pm 0.11$)\\
& Hidden Replay(ER $+$ GR) & $-9.17$ ($ \pm 2.8$) & $1.57$ ($ \pm 0.42$) & $1.21$ ($ \pm 0.33$) & $1.15$ ($ \pm 0.07$) & $0.75$ ($ \pm 0.04$) & $0.91$ ($ \pm 0.04$) & $0.79$ ($ \pm 0.02$) & $0.59$ ($ \pm 0.07$)\\
& Hidden Replay (ER $+$ RaR $+$ GR) & $-3.05$ ($ \pm 0.97$) & $1.42$ ($ \pm 0.06$) & $1.0$ ($ \pm 0.02$) & $1.17$ ($ \pm 0.06$) & $0.8$ ($ \pm 0.04$) & $0.9$ ($ \pm 0.04$) & $0.83$ ($ \pm 0.04$) & $0.59$ ($ \pm 0.06$)\\
    \hline
    \end{tabular}
    }
    \caption{Effect of different replay mechanisms during sleep on lifelong learning metrics on the \SC{} domain. Results are collected using the hidden replay model on twelve syllabi. We report means and 95\% confidence intervals. Hidden replay using only experience replay (ER) and generative replay (GR) is insufficient. Random replay (RaR) is needed to stablize the learning of the feature space, and combining all three types of replay yield the best results.}
    \label{tab:replay_mechanism_ablation_sc2}
\end{table}

\subsection{Understanding the Effect of Different Model Architectures}
\label{app:archs_ablation}

In Tables \ref{tab:algo-minigrid} and \ref{tab:algo-M121518}, we compare the different replay architectures introduced in this paper with several simple benchmark models. We observe that the hidden replay model generally outperforms the sequential and two-headed architectures as well as the simple benchmark models.

\begin{table}[H]
    \centering
    \resizebox{1.0\textwidth}{!}{
    \begin{tabular}{|c|c|c|c|c|c|c||c|c|c|c|}
    \hline
    Scenario & Agent & PM & FT & BT & RP & RR\textsubscript{$\Omega$} & RR\textsubscript{$\alpha$} & RR\textsubscript{$\sigma$} & RR\textsubscript{$\upsilon$} \\
    \hline
    & Random Policy & $0.37$ ($ \pm 0.81$) $\bullet$ & $1.06$ ($ \pm 0.02$) $\circ$ & $1.13$ ($ \pm 0.18$) $\bullet$ & $0.22$ ($ \pm 0.04$) $\circ$ & $0.09$ ($ \pm 0.02$) $\circ$ & $0.09$ ($ \pm 0.02$) $\circ$ & $0.09$ ($ \pm 0.02$) $\circ$ & $0.09$ ($ \pm 0.00$) $\circ$\\
& Baseline PPO & $-48.2$ ($ \pm 12.04$) & $4.07$ ($ \pm 0.87$) & $0.46$ ($ \pm 0.12$) & $2.17$ ($ \pm 0.61$) & $0.5$ ($ \pm 0.1$) & $0.78$ ($ \pm 0.08$) & $0.64$ ($ \pm 0.08$) & $0.16$ ($ \pm 0.02$)\\
Pairwise & Hidden Replay (ER) & $-45.25$ ($ \pm 13.54$) & $6.34$ ($ \pm 1.4$) & $0.51$ ($ \pm 0.26$) & $2.75$ ($ \pm 0.73$) & $0.47$ ($ \pm 0.06$) & $0.73$ ($ \pm 0.08$) & $0.6$ ($ \pm 0.06$) & $0.11$ ($ \pm 0.02$) $\circ$\\
& Sequential (ER + RaR + GR) & $13.58$ ($ \pm 9.06$) $\bullet$ & $4.01$ ($ \pm 1.02$) & $9.52$ ($ \pm 5.5$) $\bullet$ & $2.57$ ($ \pm 0.72$) & $0.45$ ($ \pm 0.08$) & $0.37$ ($ \pm 0.08$) $\circ$ & $0.41$ ($ \pm 0.08$) $\circ$ & $0.08$ ($ \pm 0.02$) $\circ$\\
& Two-Headed (ER + RaR + GR) & $6.73$ ($ \pm 6.19$) $\bullet$ & $4.24$ ($ \pm 0.88$) & $3.55$ ($ \pm 4.12$) $\bullet$ & $2.85$ ($ \pm 0.72$) & $0.49$ ($ \pm 0.1$) & $0.46$ ($ \pm 0.08$) $\circ$ & $0.48$ ($ \pm 0.08$) & $0.1$ ($ \pm 0.02$) $\circ$\\
& Hidden Replay (ER + RaR + GR) & $-15.77$ ($ \pm 10.09$) & $5.87$ ($ \pm 1.24$) & $1.21$ ($ \pm 0.55$) & $2.47$ ($ \pm 0.69$) & $0.6$ ($ \pm 0.06$) & $0.69$ ($ \pm 0.08$) & $0.64$ ($ \pm 0.06$) & $0.11$ ($ \pm 0.02$) $\circ$\\
\hline
& Random Policy & $-0.02$ ($ \pm 0.26$) $\bullet$ & $1.08$ ($ \pm 0.02$) $\circ$ & $1.08$ ($ \pm 0.02$) $\circ$ & $0.19$ ($ \pm 0.00$) $\circ$ & $0.09$ ($ \pm 0.00$) $\circ$ & $0.09$ ($ \pm 0.00$) $\circ$ & $0.09$ ($ \pm 0.00$) $\circ$ & $0.09$ ($ \pm 0.00$) $\circ$\\
& Baseline PPO & $-38.92$ ($ \pm 3.98$) & $4.72$ ($ \pm 0.75$) & $4.06$ ($ \pm 0.73$) & $1.44$ ($ \pm 0.2$) & $0.22$ ($ \pm 0.04$) & $0.64$ ($ \pm 0.04$) & $0.33$ ($ \pm 0.04$) & $0.15$ ($ \pm 0.02$)\\
Condensed & Hidden Replay (ER) & $-57.71$ ($ \pm 3.61$) & $6.45$ ($ \pm 0.81$) & $5.32$ ($ \pm 0.81$) & $2.04$ ($ \pm 0.1$) $\bullet$ & $0.17$ ($ \pm 0.02$) & $0.77$ ($ \pm 0.04$) $\bullet$ & $0.33$ ($ \pm 0.02$) & $0.12$ ($ \pm 0.02$)\\
& Sequential (ER + RaR + GR)& $-6.8$ ($ \pm 2.29$) $\bullet$ & $4.49$ ($ \pm 0.57$) & $1.98$ ($ \pm 0.35$) $\circ$ & $2.13$ ($ \pm 0.08$) $\bullet$ & $0.56$ ($ \pm 0.04$) $\bullet$ & $0.62$ ($ \pm 0.02$) & $0.58$ ($ \pm 0.02$) $\bullet$ & $0.12$ ($ \pm 0.02$)\\
& Two-Headed (ER + RaR + GR) & $-9.69$ ($ \pm 2.87$) $\bullet$ & $4.1$ ($ \pm 0.57$) & $1.65$ ($ \pm 0.41$) $\circ$ & $2.15$ ($ \pm 0.1$) $\bullet$ & $0.59$ ($ \pm 0.04$) $\bullet$ & $0.69$ ($ \pm 0.04$) & $0.62$ ($ \pm 0.04$) $\bullet$ & $0.13$ ($ \pm 0.02$)\\
& Hidden Replay (ER + RaR + GR) & $-21.94$ ($ \pm 2.09$) & $5.93$ ($ \pm 0.79$) & $1.32$ ($ \pm 0.2$) $\circ$ & $2.1$ ($ \pm 0.1$) $\bullet$ & $0.56$ ($ \pm 0.04$) $\bullet$ & $0.8$ ($ \pm 0.02$) $\bullet$ & $0.63$ ($ \pm 0.02$) $\bullet$ & $0.16$ ($ \pm 0.02$)\\
        \hline
    \end{tabular}
    }
    \caption{Comparison of the different model-free generative replay architectures and the baseline on two different Minigrid scenarios. The metrics are averaged over 36 lifetimes per scenario and the 95\% confidence interval is shown. $\bullet$ denotes statistically significant improvement compared to the baseline with a Kruskal-Wallis test followed by Dunn's test with Bon Ferroni correction w/ a p-value of 0.05. $\circ$ denotes the metric is statistically significantly worse than the baseline. We see that the LL agents significantly outperform the non-LL agent interms of performance maintainence and RR\textsubscript{$\Omega$}, suggesting the LL agents significantly reduce catastrophic forgetting. The hidden replay model achieves significantly higher RP and  RR\textsubscript{$\alpha$} suggesting that it enables the model to learn new tasks with minimal interference from learning previous tasks.} 
    \label{tab:algo-minigrid}
\end{table}

\begin{table}[H]
    \centering
    \resizebox{1.0\textwidth}{!}{
    \begin{tabular}{|c|c|c|c|c|c||c|c|c|c|}
    \hline
    Scenario & Agent & PM & FT & BT & RP & RR\textsubscript{$\Omega$} & RR\textsubscript{$\alpha$} & RR\textsubscript{$\sigma$} & RR\textsubscript{$\upsilon$} \\
    \hline
& Random Policy & $0.0$ ($ \pm 0.00$) & $1.0$ ($ \pm 0.00$) & $1.0$ ($ \pm 0.00$) & $0.56$ ($ \pm 0.04$) $\circ$ & $0.33$ ($ \pm 0.06$) $\circ$ & $0.33$ ($ \pm 0.06$) $\circ$ & $0.33$ ($ \pm 0.06$) $\circ$ & $0.34$ ($ \pm 0.14$)\\
& Baseline VTrace (No Sleep) & $-8.2$ ($ \pm 6.54$) & $1.11$ ($ \pm 0.23$) & $0.85$ ($ \pm 0.21$) & $1.03$ ($ \pm 0.06$) & $0.74$ ($ \pm 0.14$) & $0.82$ ($ \pm 0.12$) & $0.78$ ($ \pm 0.12$) & $0.39$ ($ \pm 0.17$)\\
Pairwise & Sequential (ER $+$ GR)& $-5.51$ ($ \pm 3.41$) & $1.18$ ($ \pm 0.36$) & $0.77$ ($ \pm 0.13$) & $0.89$ ($ \pm 0.06$) & $0.54$ ($ \pm 0.15$) & $0.61$ ($ \pm 0.15$) & $0.57$ ($ \pm 0.15$) & $0.47$ ($ \pm 0.24$)\\
& Two-Headed (ER $+$ GR)& $-2.35$ ($ \pm 7.31$) & $1.46$ ($ \pm 0.48$) & $0.98$ ($ \pm 0.18$) & $0.95$ ($ \pm 0.11$) & $0.76$ ($ \pm 0.18$) & $0.79$ ($ \pm 0.18$) & $0.77$ ($ \pm 0.18$) & $0.41$ ($ \pm 0.2$)\\
& Hidden Replay (ER Only) & $-7.9$ ($ \pm 5.11$) & $1.14$ ($ \pm 0.4$) & $0.75$ ($ \pm 0.29$) & $0.92$ ($ \pm 0.09$) & $0.87$ ($ \pm 0.11$) & $0.9$ ($ \pm 0.11$) & $0.8$ ($ \pm 0.13$) & $0.37$ ($ \pm 0.2$)\\
& Hidden Replay (ER $+$ RaR $+$ GR) & $-0.57$ ($ \pm 2.93$) & $1.85$ ($ \pm 0.88$) & $0.95$ ($ \pm 0.15$) & $1.04$ ($ \pm 0.11$) & $0.77$ ($ \pm 0.18$) & $0.82$ ($ \pm 0.13$) & $0.8$ ($ \pm 0.15$) & $0.55$ ($ \pm 0.26$)\\
\hline
& Random Policy & $0.0$ ($ \pm 0.00$) $\bullet$ & $1.0$ ($ \pm 0.00$) & $1.0$ ($ \pm 0.00$) $\bullet$ & $0.42$ ($ \pm 0.04$) $\circ$ & $0.32$ ($ \pm 0.09$) $\circ$ & $0.32$ ($ \pm 0.09$) $\circ$ & $0.32$ ($ \pm 0.09$) $\circ$ & $0.32$ ($ \pm 0.22$)\\
& Baseline VTrace (No Sleep) & $-8.99$ ($ \pm 3.06$) & $1.11$ ($ \pm 0.23$) & $0.79$ ($ \pm 0.12$) & $0.92$ ($ \pm 0.04$) & $0.9$ ($ \pm 0.14$) & $0.91$ ($ \pm 0.12$) & $0.82$ ($ \pm 0.12$) & $0.39$ ($ \pm 0.17$)\\
Alternating & Sequential (ER $+$ GR)& $-7.44$ ($ \pm 3.43$) & $1.18$ ($ \pm 0.36$) & $0.88$ ($ \pm 0.09$) & $0.8$ ($ \pm 0.09$) & $0.59$ ($ \pm 0.15$) & $0.66$ ($ \pm 0.13$) & $0.6$ ($ \pm 0.13$) & $0.47$ ($ \pm 0.24$)\\
& Two-Headed (ER $+$ GR)& $-7.7$ ($ \pm 4.89$) & $1.46$ ($ \pm 0.48$) & $0.91$ ($ \pm 0.11$) & $0.9$ ($ \pm 0.09$) & $0.88$ ($ \pm 0.22$) & $0.94$ ($ \pm 0.11$) & $0.84$ ($ \pm 0.15$) & $0.41$ ($ \pm 0.2$)\\
& Hidden Replay (ER Only) & $-7.9$ ($ \pm 5.11$) & $1.14$ ($ \pm 0.4$) & $0.75$ ($ \pm 0.29$) & $0.92$ ($ \pm 0.09$) & $0.87$ ($ \pm 0.11$) & $0.9$ ($ \pm 0.11$) & $0.8$ ($ \pm 0.13$) & $0.37$ ($ \pm 0.2$)\\
& Hidden Replay (ER $+$ RaR $+$ GR) & $-6.13$ ($ \pm 4.64$) & $1.85$ ($ \pm 0.88$) & $0.87$ ($ \pm 0.13$) & $0.91$ ($ \pm 0.09$) & $0.86$ ($ \pm 0.13$) & $0.89$ ($ \pm 0.13$) & $0.82$ ($ \pm 0.15$) & $0.55$ ($ \pm 0.26$)\\
\hline
& Random & $-0.03$ ($ \pm 0.28$) $\bullet$ & $1.01$ ($ \pm 0.02$) & $1.01$ ($ \pm 0.02$) & $0.54$ ($ \pm 0.02$) $\circ$ & $0.31$ ($ \pm 0.02$) $\circ$ & $0.3$ ($ \pm 0.02$) $\circ$ & $0.31$ ($ \pm 0.00$) $\circ$ & $0.3$ ($ \pm 0.04$) $\circ$\\
& Baseline VTrace (No Sleep) & $-3.41$ ($ \pm 1.7$) & $1.19$ ($ \pm 0.08$) & $1.17$ ($ \pm 0.29$) & $1.14$ ($ \pm 0.06$) & $0.64$ ($ \pm 0.06$) & $0.74$ ($ \pm 0.06$) & $0.66$ ($ \pm 0.06$) & $0.49$ ($ \pm 0.06$)\\
Condensed & Sequential (ER $+$ GR) & $-3.7$ ($ \pm 1.58$) & $1.15$ ($ \pm 0.04$) & $1.0$ ($ \pm 0.09$) & $0.91$ ($ \pm 0.09$) $\circ$ & $0.53$ ($ \pm 0.13$) & $0.69$ ($ \pm 0.09$) & $0.61$ ($ \pm 0.09$) & $0.49$ ($ \pm 0.11$)\\
& Two-Headed (ER $+$ GR) & $-4.53$ ($ \pm 2.97$) & $1.34$ ($ \pm 0.09$) & $1.15$ ($ \pm 0.11$) & $1.09$ ($ \pm 0.13$) & $0.76$ ($ \pm 0.11$) & $0.87$ ($ \pm 0.09$) & $0.78$ ($ \pm 0.09$) & $0.57$ ($ \pm 0.09$)\\
& Hidden Replay (ER Only) & $-8.65$ ($ \pm 3.15$) $\circ$ & $1.42$ ($ \pm 0.15$) $\bullet$ & $1.12$ ($ \pm 0.11$) & $1.12$ ($ \pm 0.09$) & $0.73$ ($ \pm 0.09$) & $0.89$ ($ \pm 0.07$) & $0.77$ ($ \pm 0.04$) & $0.55$ ($ \pm 0.07$)\\
& Hidden Replay (ER $+$ RaR $+$ GR) & $-3.05$ ($ \pm 0.97$) & $1.42$ ($ \pm 0.06$) $\bullet$ & $1.0$ ($ \pm 0.02$) & $1.17$ ($ \pm 0.06$) & $0.8$ ($ \pm 0.04$) & $0.9$ ($ \pm 0.04$) & $0.83$ ($ \pm 0.04$) $\bullet$ & $0.59$ ($ \pm 0.06$)\\
    \hline
    \end{tabular}
    }
    \caption{Comparison of the different model-free generative replay architectures and the baseline on three different \SC{} scenarios. The metrics are averaged over 12 lifetimes per scenario and the 95\% confidence interval is shown. $\bullet$ denotes statistically significant improvement compared to the baseline with a Kruskal-Wallis test followed by Dunn's test with Bon Ferroni correction w/ a p-value of 0.05. $\circ$ denotes the metric is statistically significantly worse than the baseline. In general, the hidden replay model performs very well, showing significant gains in FT. Interestingly, the baseline agent performs well in learning new tasks, but the continual learners exhibit better performance on unseen tasks, indicating that they exhibit better generalization. RR\textsubscript{$\Omega$} also shows the continual learners are able to learn without forgetting.}
    \label{tab:algo-M121518}
\end{table}

\subsection{Qualitative Analysis of the Generative Model}
\label{sec:reconstructions_app}

\begin{figure}[H]
\centering
\includegraphics[width=0.95\textwidth]{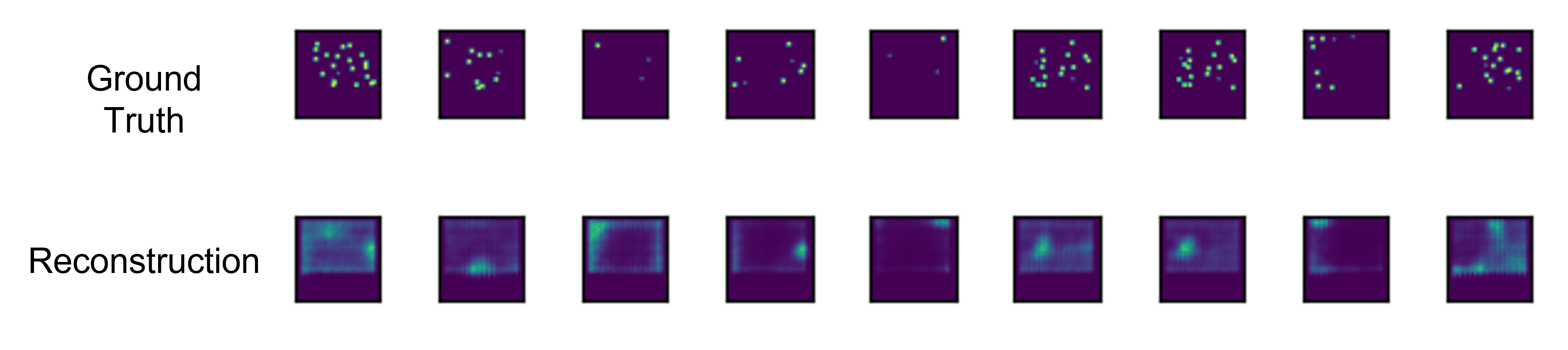}
\caption{Reconstructions of the first channel of the ``CollectMineralShards'' SC-2 task using the sequential architecture after 11,000 sleep batches, imitating a converged wake model.}
\label{fig:recon_sequential_collect}
\end{figure}

\begin{figure}[H]
\centering
\includegraphics[width=0.95\textwidth]{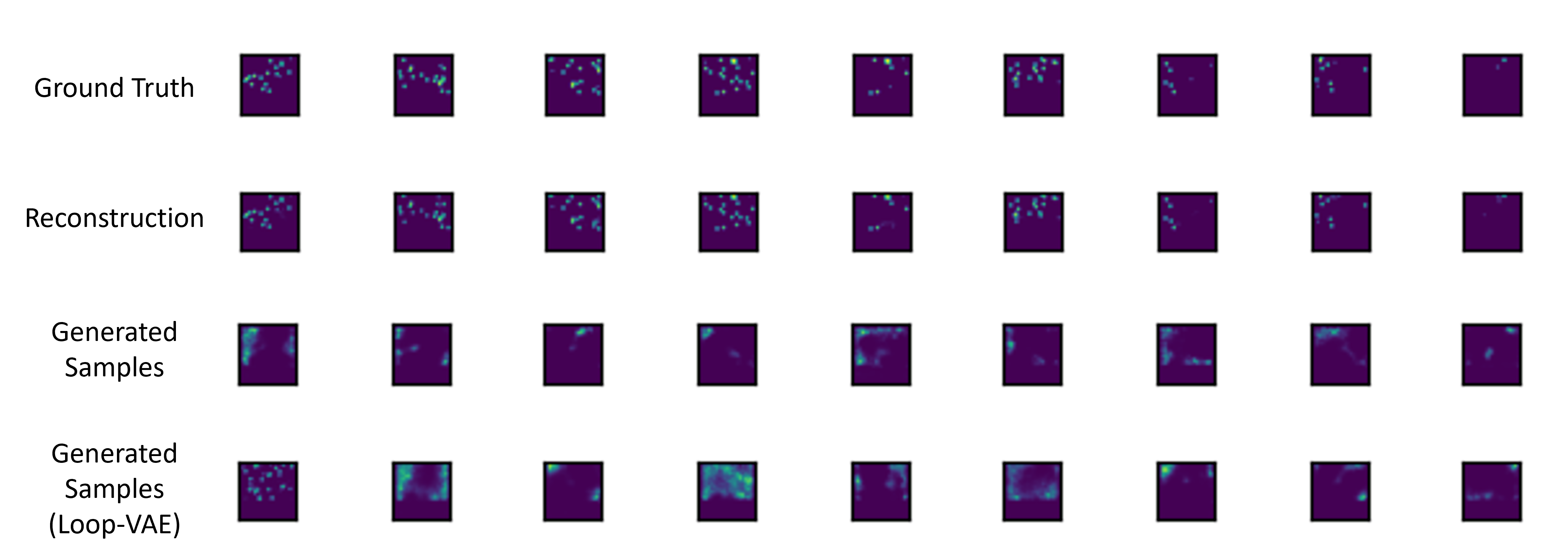}
\caption{Reconstructions of the first channel of the ``CollectMineralShards'' SC-2 task using the two-headed architecture after 11,000 sleep batches, imitating a converged wake model. We also show some generated samples and the effect of passing generating via the ``loop-vae'' procedure.}
\label{fig:recon_two_headed_collect}
\end{figure}

\begin{figure}[H]
\centering
\includegraphics[width=0.95\textwidth]{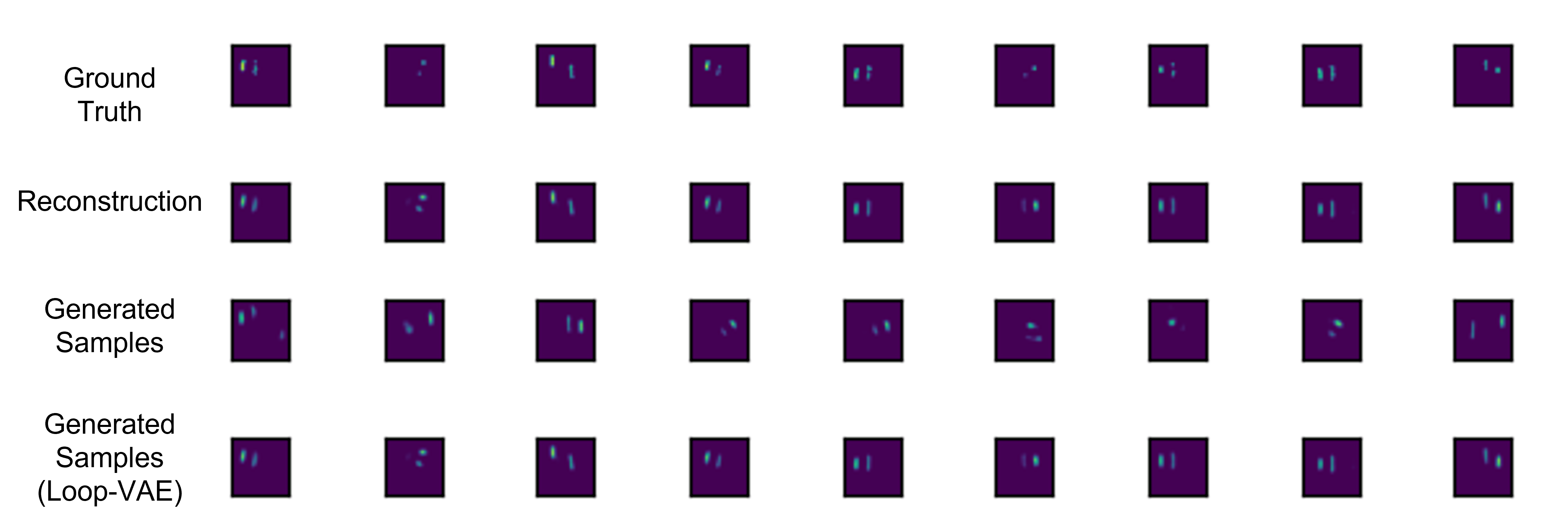}
\caption{Reconstructions and generated samples of the first channel of the ``DefeatRoaches'' SC-2 task using the two-headed architecture.}
\label{fig:recon_two_headed_defeat}
\end{figure}

In this section, we qualitatively evaluate the capabilities of each replay architecture with respect to its generative modeling capabilities. 
In Fig.~\ref{fig:recon_sequential_collect}, we show reconstructions of the first channel of the ``CollectMineralShards'' SC-2 task using the sequential architecture after 11K sleep batches, imitating a converged wake model. These observations are challenging to reconstruct and generate due to the random distribution of shards, each represented by a few pixels. The sequential architecture produces blurry reconstructions that resemble density maps over the shard locations. In Fig.~\ref{fig:recon_two_headed_collect}, we show reconstructions on the same task for the two-headed architecture. By separating the generative model from the policy network, the model reconstructs the observations with sharper detail; however, the two-headed model still struggles with generating realistic observations, generating observations that degrade to density maps. While the generated observations are sub-optimal in terms of realism, experiments suggest they still provide useful information for training the policy network. We show additional ground-truth, reconstructions, and generated samples for the ``DefeatRoaches'' task using the two-headed architecture in Fig.~\ref{fig:recon_two_headed_defeat}. In this case, there is much more structural regularity in the observations, so the reconstructions and generated samples better resemble ground truth data.


\begin{figure}[H]
\centering
\includegraphics[width=0.7\textwidth]{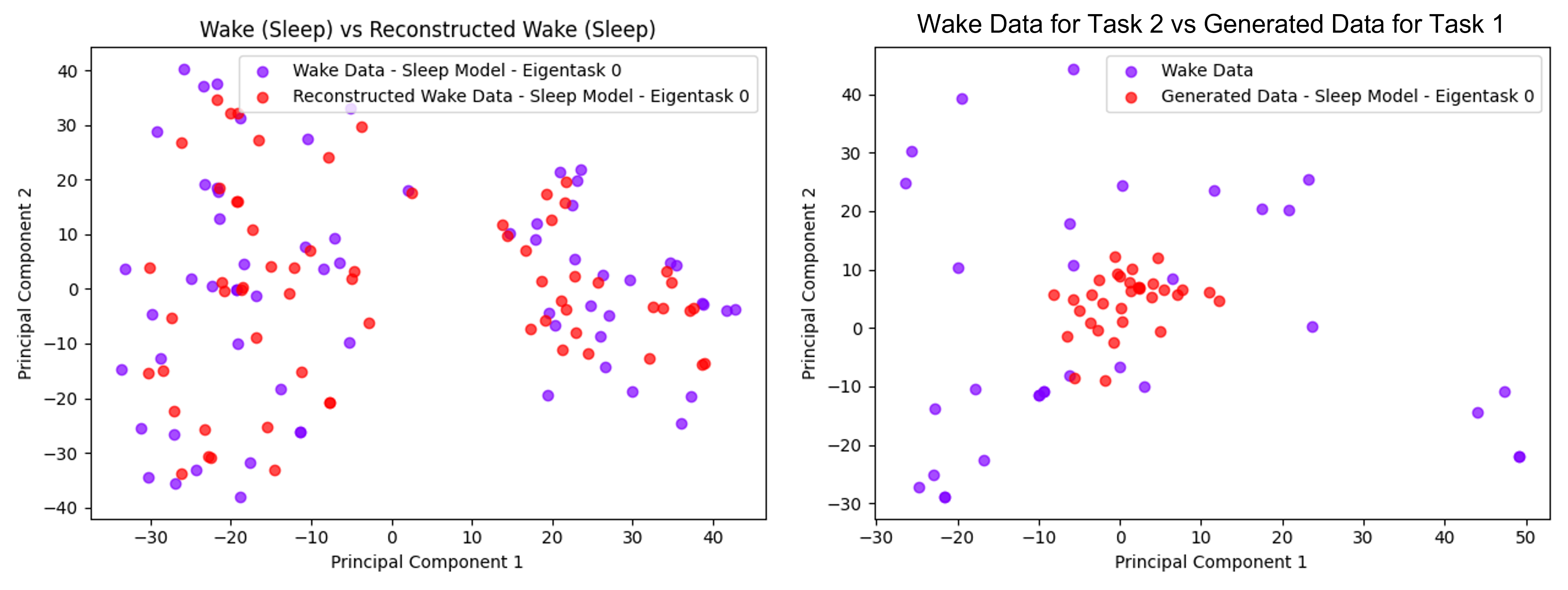}
\caption{To evaluate reconstructions and generated samples for the hidden replay architecture, we project data in the feature space via its principle components. Left: Reconstructions of the features for the ``DefeatRoaches'' task. Right: Overlaying the feature distribution of ``DefeatZerglingsAndBanelings'' and the generated samples for ``DefeatRoaches'' after learning the task sequence ``DefeatRoaches'' $ \rightarrow$ ``DefeatZerglingsAndBanelings''.}
\label{fig:recon_hidden_replay}
\end{figure}

Visualizing the reconstructions and generated samples of the hidden replay model is more challenging. To do so, we use a two-dimensional principle components analysis to compare ``ground-truth'' features (i.e., the output of the feature extractor) with reconstructions and generated features. On the left side of Fig.\ \ref{fig:recon_hidden_replay}, the VAE is able to closely reconstruct the ground-truth features for the ``DefeatRoaches'' task. On the right side, we overlay the feature distribution of ``DefeatZerglingsAndBanelings'' and the generated samples for ``DefeatRoaches'' after learning the task sequence ``DefeatRoaches'' $ \rightarrow$ ``DefeatZerglingsAndBanelings''. Since these tasks are similar, we see an overlap between the generated features of ``DefeatRoaches'' from the previous sleep with the new features of ``DefeatZerglingsAndBanelings'', suggesting the generator is generating meaningful features from the task seen during the previous sleep phase.

\subsection{Additional Results: Understanding the Interplay Between Different Replay Mechanisms}
\label{sec:justification_of_exemplars}

\begin{figure}[H]
\centering
\includegraphics[width=0.8\textwidth]{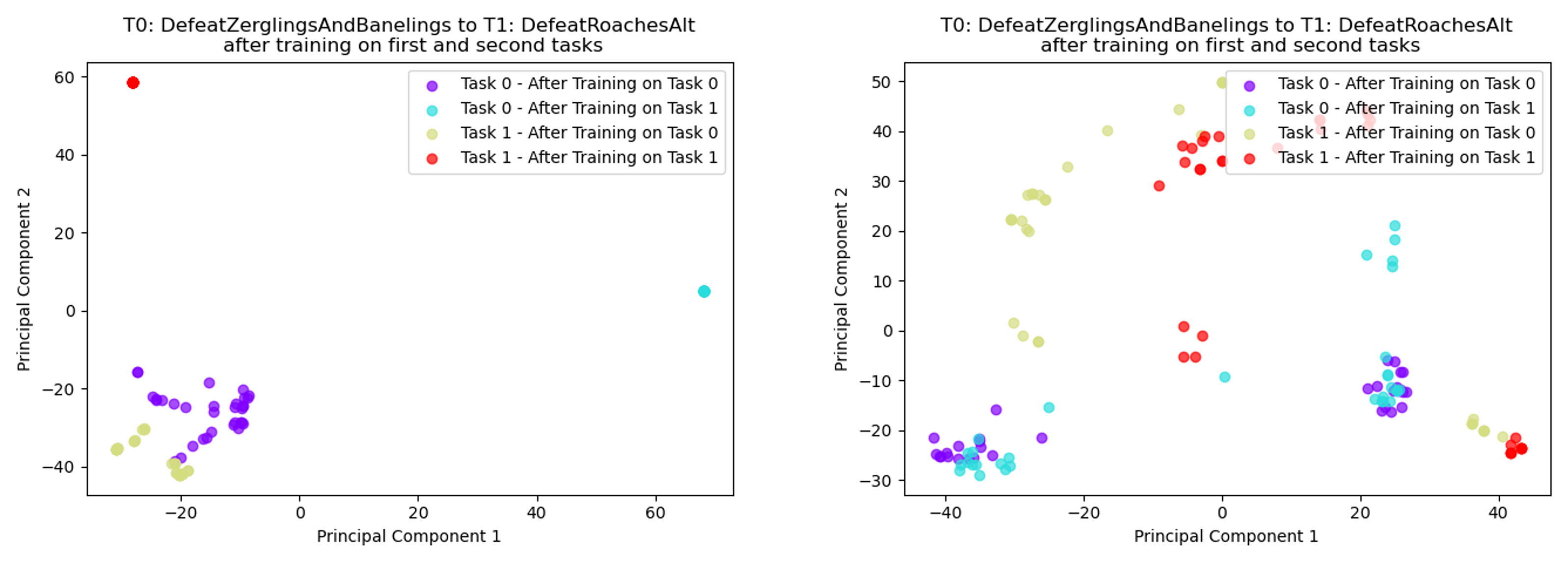}
\caption{We show the necessity of combining hidden replay with random replay. Left: Using hidden replay without random replay results in the feature space for task 0 dramatically changing after training on task 1, and the feature space for both classes converge to a point. PM is -10.68. Right: The distribution of task 0 before and after training on task 1 are significantly better aligned when hidden replay is used in conjunction with random replay. PM is -1.93.}
\label{fig:exemplar_replay_1}
\end{figure}

\begin{figure}[H]
\centering
\includegraphics[width=0.95\textwidth]{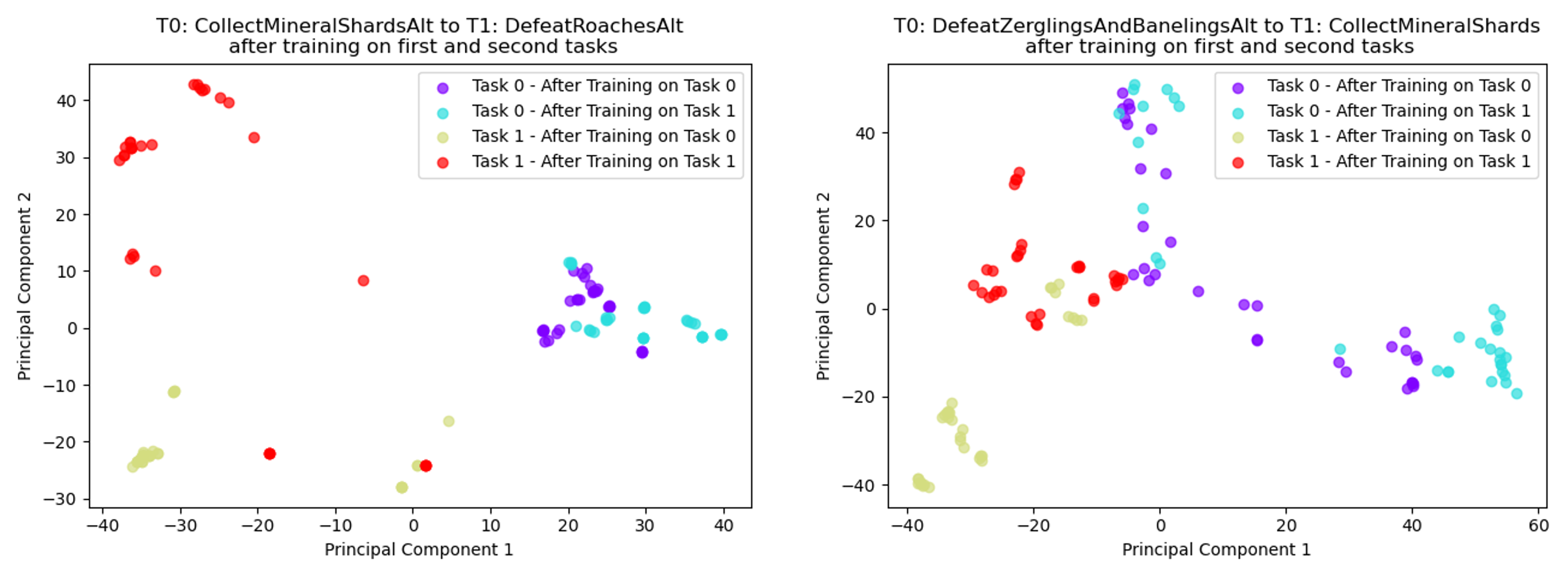}
\caption{We show two examples where hidden + random replay maintains the feature space of the previous task after learning a new dissimilar SC-2 task (``Collect'' $\rightarrow$ ``Defeat'' and vice versa).}
\label{fig:exemplar_replay_2}
\end{figure}

We've observed that generative replay alone is not sufficient for the hidden replay architecture. This is because there is no constraint limiting the drift in the feature space between sleeps, so the feature extractor can completely change between two time steps if the tasks are perceptually dissimilar or dissimilar in policy. To help to overcome this issue, we employ ``random replay'' where we save a small batch of random observation-action pairs in the original observation space after every sleep, and randomly replay them during every sleep.  In Fig.\ \ref{fig:exemplar_replay_1}, we show the necessity of combining hidden replay with random replay. On the left, we see that using hidden replay without random replay results in the feature space for task 0 dramatically changing after training on task 1, and PM is low. On the right, we see that the distribution of task 0 before and after training on task 1 is significantly better aligned when hidden replay is used in conjunction with random replay, and PM significantly improves. Additional visualizations for pairs of dissimilar tasks are seen in Fig.~\ref{fig:exemplar_replay_2}. In Table \ref{tab:exemplars_vs_no_exemplars}, we see using random replay generally improves PM and FTR, which agrees with our hypothesis that it helps to maintain the feature space, leading to less forgetting and potentially features that generalize better to other tasks. These results are inline with similar observations made from the main experiments of this paper.

\begin{table}[H]
    \centering
    \begin{tabular}{|c|c|c|c|c|c|}
        \hline
         Scenario & Agent & PM & FT & BT & RP \\
         \hline
          Pairwise & No Random Replay & $-6.28$ ($ \pm 12.82$) & $1.35$ ($ \pm 0.24$) & $0.80$ ($ \pm 0.37$) & $1.08$ ($ \pm 0.19$) \\
         & Random Replay & $-3.03$ ($ \pm 4.92$) & $1.65$ ($ \pm 0.28$) & $0.92$ ($ \pm 0.13$) & $1.04$ ($ \pm 0.17$) \\
         \hline
        Condensed & No Random Replay & $-9.35$ ($ \pm 3.69$) & $1.37$ ($ \pm 0.15$) & $1.20$ ($ \pm 0.24$) & $1.16$ ($ \pm 0.16$) \\
         & Random Replay & $-3.05$ ($ \pm 1.76$) & $1.42$ ($ \pm 0.11$) & $1.00$ ($ \pm 0.03$) & $1.17$ ($ \pm 0.11$) \\
         \hline
    \end{tabular}
    \caption{Comparison of the hidden replay model with and without random replay. Results are averaged over multiple lifetimes, and we also record the standard deviation of the metrics.}
    \label{tab:exemplars_vs_no_exemplars}
\end{table}

\end{document}